\documentclass[acmsmall]{acmart}

\AtBeginDocument{%
  }

\setcopyright{acmcopyright}
\copyrightyear{0000}
\acmYear{0000}
\acmDOI{000000}

\acmJournal{JACM}
\acmVolume{00}
\acmNumber{0}
\acmArticle{000}
\acmMonth{0}

\usepackage{algorithm2e}
\usepackage{amsmath}

\begin{document}

\title{Mastering the exploration-exploitation trade-off in Bayesian Optimization}

\author{Antonio Candelieri}
\authornote{Authour note}
\email{antonio.candelieri@unimib.it}
\orcid{0000-0003-1431-576X}
\affiliation{
  \institution{University of Milano-Bicocca}
  \streetaddress{Piazza dell'Ateneo Nuovo 1}
  \city{Milano}
  \country{Italy}
  \postcode{20126}
}

\renewcommand{\shortauthors}{Candelieri}

\begin{abstract}
Gaussian Process based Bayesian Optimization is a well-known sample efficient sequential strategy for globally optimizing black-box, expensive, and multi-extremal functions. The role of the Gaussian Process is to provide a probabilistic approximation of the unknown function, depending on the sequentially collected observations, while an \textit{acquisition function} drives the choice of the next solution to evaluate, balancing between \textit{exploration} and \textit{exploitation}, depending on the current Gaussian Process model. Despite the huge effort of the scientific community in defining effective exploration-exploitation mechanisms, we are still far away from \textit{the master acquisition function}. This paper merges the most relevant results and insights from both algorithmic and human search strategies to propose a novel acquisition function, mastering the trade-off between explorative and exploitative choices, adaptively. We compare the proposed acquisition function on a number of test functions and against different state-of-the-art ones, which are instead based on prefixed or random scheduling between exploration and exploitation. A Pareto analysis is performed with respect to two (antagonistic) goals: convergence to the optimum and exploration capability. Results empirically prove that the proposed acquisition function is almost always Pareto optimal and also the most balanced trade-off between the two goals.
\end{abstract}

\ccsdesc[500]{Computing methodologies~Search methodologies}
\ccsdesc[500]{Computing methodologies~Continuous space search}
\ccsdesc[500]{Computing methodologies~Machine learning}

\keywords{Bayesian Optimization, Gaussian Process, Acquisition Function, Confidence Bound}

\received{xxx}
\received[revised]{xxx}
\received[accepted]{xxx}

\maketitle

\section{Introduction}
Bayesian Optimization (BO)~\cite{frazier2018bayesian,archetti2019bayesian,garnett2023bayesian} is the most widely adopted method for globally optimizing black-box, expensive, and multi-extremal functions. Thanks to its sample efficiency, it is has been successfully applied to an extremely wide range of real-life applications, such as engineering~\cite{lam2018advances,wang2021nextorch,wang2022bayesian}, optimal control of complex systems~\cite{baheri2017real,candelieri2020learning,candelieri2023safe,candelieri2018bayesian}, and Automated Machine Learning~\cite{zoller2021benchmark,waring2020automated,hutter2019automated,perego2022autotinyml,candelieri2021green,candelieri2022fair}.

The basic BO algorithm is a \textit{sequential learning-and-optimization framework}, based on two components: a \textit{probabilistic surrogate model}, approximating the objective function depending on all the solutions evaluated so far (i.e., the \textit{learning} component of BO), and an \textit{acquisition function} (aka \textit{infill criterion} or \textit{utility function}), driving the choice of the next solution to evaluate (i.e., the \textit{optimization} component), depending on the current probabilistic surrogate model. The basic BO algorithm is schematically reported in \textbf{Algorithm~\ref{algo:bo}}.

\RestyleAlgo{ruled}

\begin{algorithm}
\caption{Basic Bayesian Optimization algorithm}\label{algo:bo}
\KwData{\\ \quad$\Omega\subset\mathbb{R}^d$, the usually box-bounded Search Space;
\\ \quad$\mathcal{D}=\big\{\big(\mathbf{x}^{(i)},y^{(i)}\big)\big\}$, an initial set of $n$ observations, with $\mathbf{x}^{(i)},\dots,\mathbf{x}^{(n)}$ usually from Latin\\ \quad\quad Hypercube Sampling (LHS) and $y^{(i)}=f\big(\mathbf{x}^{(i)}\big)+\varepsilon^{(i)}$, $\varepsilon^{(i)}\sim \mathcal{N}(0,\lambda^2)$;
\\ \quad$N>n$.}
 \While{$n<N$}{
  fit the probabilistic surrogate model $\mathcal{M}$ on $\mathcal{D}$\;
  obtain $\mathbf{x}^{(n+1)}$ as the optimizer of the acquisition function $\mathcal{A}_\mathcal{M}(\mathbf{x})$\;
  evaluate $\mathbf{x}^{(n+1)}$ and observe $y^{(n+1)}=f\big(\mathbf{x}^{(i)}\big)+\varepsilon^{(i)}$\;
  $\mathcal{D} \leftarrow \mathcal{D} \cup \big\{\big(\mathbf{x}^{(n+1)},y^{(n+1)}\big)\big\}$\;  
  $n \leftarrow n+1$
 }
 \KwResult{\\ \quad$(\mathbf{x}^+,y^+): y^+=\min\big\{y^{(i)}:\big(\mathbf{x}^{(i)},y^{(i)}\big) \in \mathcal{D}\big\}$ (``max'' in the case of a maximization problem.)} 
\end{algorithm}

\vspace{-0.2cm}

During the last decade, research has focused on two topics: (\textit{a}) extending the set of modelling techniques so that the most appropriate probabilistic surrogate model can be chosen with respect to the specific features of the target optimization problem, and (\textit{b}) defining acquisition functions to effectively deal with the \textit{exploration-exploitation dilemma}. This paper focuses on the second topic only, and consider the Gaussian Process (GP) regression~\cite{gramacy2020surrogates,williams2006gaussian} as probabilistic modelling technique. This is not a limitation because the probabilistic surrogate model and the acquisition function can be independently chosen, without affecting the overall BO framework, as proven by all the widely known BO software libraries~\cite{bischl2017mlrmbo,balandat2020botorch,lindauer2022smac3,dewancker2016bayesian,golovin2017google}. Moreover, although GP regression is well suited for modelling functions over continuous search spaces, extensions have been recently proposed to also deal with more complicated ones~\cite{ru2020bayesian,nguyen2020bayesian,cuesta2022comparison}.\\

As far as the acquisition function is concerned, the alternatives proposed over time can be conveniently divided into \textit{improvement-based} and \textit{information-based} acquisition functions~\cite{shahriari2015taking}. While the former are devoted to search for the \textit{optimum} (i.e., the best objective function's value), the latter search for the \textit{optimizer} (i.e., the solution providing to the best objective function's value). Although apparently irrelevant, this distinction leads to completely different strategies, with information-based acquisition functions usually offering a greater sample efficiency, but entailing a larger computational cost to determine the next solution to evaluate. From a theoretical perspective, it is important to quote that a very recent work~\cite{neiswanger2022generalizing} shows that popular acquisition functions -- improvement as well as information based -- are just special cases of a generalization of the Shannon entropy, from statistical decision theory.

This paper considers the improvement-based acquisition functions only, which combine the model's prediction and the associated predictive uncertainty to implement their own exploration-exploitation trade-off mechanisms. The common underlying consists into avoiding decisions exclusively based on the model's prediction, because this would exclusively lead to \textit{local search}, with the risk to remain trapped into local optima. Thus, every improvement-based acquisition function includes an \textit{uncertainty bonus} to guarantee \textit{global search} and, consequently, convergence to the global optimum.
The most recent research studies have proposed \textit{prefixed} (usually increasing) or \textit{random scheduling} of the uncertainty bonus. However, results are inconclusive, with every proposed approach beating the previous ones only on a subset of test problems, and often entailing one or more hyperparameters to be manually fine-tuned, for every specific problem.\\

\noindent
\textbf{Contribution.} Here we summarize the main contribution of this paper. First, we have identified the main limitations of the current improvement-based acquisition functions:
\begin{itemize}
    \item both prefixed and random scheduling do not consider what is the \textit{impact} of every sequential decision on the model and, consequently, the decision suggested by the acquisition function. On the contrary, we prove that this information is crucial for the successive decision;
    \item as already proven in~\cite{de2021greed,vzilinskas2019bi}, the decisions generated by improvement-based acquisition functions are \textit{Pareto rational} with respect to two objectives: optimizing according to the model's prediction (i.e., exploitation) and maximizing the predictive uncertainty (i.e., exploration). Varying the uncertainty bonus translates into move on the Pareto front only;
    \item on the contrary, \textit{human-search} is characterized by the chance to also make non-Pareto rational decisions, with respect to the same two objectives~\cite{candelieri2021uncertainty,candelieri2022explaining}. Moreover, the number of non-Pareto rational decisions does not follow any prefixed or random scheduling, but it seems to be triggered by the impact of every decision;
    \item improvement-based acquisition functions rely on the predictive uncertainty of the probabilistic surrogate model. However, it is biased by the prediction itself, especially when GP regression is used. Even more critical, \textit{variance starvation} (i.e., predictive uncertainty close to zero) may occur with GP regression, nullifying any uncertainty bonus.\\   
\end{itemize}

As a consequence of these considerations, we propose a novel acquisition function that is able \textit{\textbf{to adaptively provide exploitative or explorative decisions, depending on the evolution of the sequential optimization process, instead of using a random or prefixed scheduling for uncertainty bonus}}.\\

More specifically, our master acquisition function:
\begin{itemize}
    \item uses the prediction of the GP model, only, to make its decision about the next solution to evaluate (pure exploitative decision);
    \item discards that decision if it has a small chance to significantly improve the current best solution (as explained in Section~\ref{sec:3}) and, instead, makes a different decision aimed at \textit{uncertainty reduction}. To do this, we do not use the GP's predictive uncertainty but a different uncertainty quantification measure, namely the Inverse Distance Weighting (also introduced in Section~\ref{sec:3}), which is not biased by the GP's prediction.\\
\end{itemize}

In this paper, a number of state-of-the-art acquisition functions have been considered and applied on a number of test problems. We have proposed a Pareto analysis of the effectiveness of these acquisition functions, based on two metrics: convergence to the optimum and capacity to explore. This could become a more mathematically sound and general method to compare the acquisition functions for BO.

Finally, it is important to remark that the aim of this paper is not to be a survey or a review on the topic.

\textbf{Related works.} Here we want to summarize the most related works, at our knowledge, and their main contributions, achievements, and limitations.
In \cite{srinivas2012information} the first proof of convergence of the well-known GP Confidence Bound (GP-CB) acquisition function is provided. Basically, the uncertainty bonus is logarithmically increased with the number of function evaluations in order to increase exploration and reduce the chance to get stuck into local optima. More recently, \cite{berk2020randomised} has empirically proved that a random scheduling of the bonus uncertainty (i.e., sampling from a Gamma distribution whose parameters are adjusted depending on the number of function evaluations) can outperform the previous scheduling on an number of test problems. In \cite{vzilinskas2019bi}, a Pareto analysis of every decision implied by traditional acquisition functions is performed, where the two objectives are the prediction of the GP model and the associated predictive uncertainty. As a result, varying the value of the uncertainty bonus translates into moving onto the Pareto front: thus, GP-CB only allows for Pareto decisions, whichever is the scheduling of the uncertainty bonus. According to this, \cite{de2021greed} proposed a new mechanism based on the $\varepsilon$-greedy strategy typically adopted in Reinforcement Learning. With probability $(1-\varepsilon)$ a \textit{greedy} decision is taken (i.e., the solution optimizing the GP's prediction), while a random choice is performed with probability $\varepsilon$. Two alternatives are possible in this case: randomly selecting among Pareto rational decisions or uniformly at random over the entire Search Space. This is interesting because in~\cite{candelieri2021uncertainty,candelieri2022explaining} it has been empirically proved that humans usually perform non-Pareto rational decision when the chance to improve is small. Thus, while the $\varepsilon$-greedy method relies on the probability $\varepsilon$ -- that is a hyperparameter to set before the optimization -- humans can adaptively switch between Pareto and non-Pareto decisions along the optimization process.
Finally, we would like to also quote a very recent work considering the alternation between two different acquisition functions, that is Expected Improvement (EI) and Probability of Improvement (PI). Indeed, these two acquisition functions offer two different exploration-exploitation trade-off, with PI more biased towards exploitation. More precisely,~\cite{benjamins2022pi} propose two schemes: alternating between PI and EI (aka round-robin) or starting with EI and then switching to PI, after a given number of iterations (i.e., a hyperparameter to be set before the optimization process). Years before, the dynamic selection of the acquisition function from a portfolio of possible choices was proposed and investigated in~\cite{hoffman2011portfolio}.\\  

\textbf{Unrelated works.} As already mentioned, the research on acquisition functions for BO is quite impressively large. It is important to remark that -- even if valuable for their contributions -- the following works are not directly related to this paper.
All the information-based acquisition functions are not related, specifically Entropy Search (ES)~\cite{hennig2012entropy}, Predictive Entropy Search (PES)~\cite{hernandez2014predictive}, Max-value Entropy Search (MES)~\cite{wang2017max}, and~\cite{hvarfner2022joint}. These methods are based on sampling from GP posterior -- usually via Thompson Sampling -- that is the operation leading to their expensive computational cost. The issue of efficiently sampling from GP posterior has been recently addressed in~\cite{wilson2020efficiently}.
The so-called Exploration Enhanced Expected Improvement (EEEI) acquisition function \cite{berk2019exploration} is also based on sampling from posterior and, therefore, it is not related to our work.
Moreover, all the look-ahead acquisitions, including Knowledge Gradient (KG) \cite{frazier2009knowledge}, are also out of scope. Finally, the trust-region based approaches, such as  TRIKE~\cite{regis2016trust}, TREGO~\cite{diouane2022trego}, and TuRBO~\cite{eriksson2019scalable} are not related to our method.

\section{Background}

\subsection{GP-based BO}
Assume to be at a generic iteration $n$, all the observations collected so far are stored into the dataset $\mathcal{D}=\big\{\mathbf{x}^{(i)},y^{(i)}\big\}$, with $\vert\mathcal{D}\vert=n$. For simplifying the notation in the following, we rewrite the dataset as $\mathcal{D}=\big\{\mathbf{X},\mathbf{y}\big\}$, where $\mathbf{X}=\big\{\mathbf{x}^{(1)},\dots,\mathbf{x}^{(n)}\big\}$ and $\mathbf{y} = \big(y^{(1)},\dots,y^{(n)}\big)$. Recall from \textbf{Algoritm~\ref{algo:bo}} that $y^{(i)}=f\big(\mathbf{x}^{(i)}\big)+\varepsilon^{(i)}$, with $f:\mathbb{R}^d\rightarrow\mathbb{R}$ the objective function, possibly noisy (i.e., $\varepsilon^{(i)}\sim\mathcal{N}(0,\lambda^2)$).

A GP regression model is used to approximate the black-box, expensive, and multiextremal objective function, depending on the dataset $\mathcal{D}$. GP regression is a well-known \textit{kernel method}~\cite{scholkopf2002learning,hofmann2008kernel}, in which covariance between observations is codified through a \textit{kernel function}, $k:\mathbb{R}^d \times \mathbb{R}^d \rightarrow \mathbb{R}$, which can be chosen among many alternatives~\cite{williams2006gaussian,gramacy2020surrogates}. In this paper we consider the Squared Exponential (SE) kernel, $k(\mathbf{x},\mathbf{x}')=\sigma_f^2 e^{-\frac{||\mathbf{x}-\mathbf{x}'||^2}{2\ell^2}}$, which assumes the smoothest approximation for the objective function. It is usually a reasonable choice in the case of experimental settings like that considered in this paper, while other \textit{less smooth} kernels can result more appropriate when addressing real-life problems.

A GP regression model is also said \textit{probabilistic}, meaning that for any given input it provides both a prediction and the associated (predictive) uncertainty, respectively given by:

\begin{equation}
    \label{eq:mu}
    \mu(\mathbf{x}) = \mathbf{k}(\mathbf{x},\mathbf{X}) \left[\mathbf{K}+\lambda^2\mathbf{I}\right]^{-1} \mathbf{y}
\end{equation}

\begin{equation}
    \label{eq:s2}
    \sigma^2(\mathbf{x}) = k(\mathbf{x},\mathbf{x}) - \mathbf{k}(\mathbf{x},\mathbf{X}) \left[\mathbf{K}+\lambda^2\mathbf{I}\right]^{-1} \mathbf{k}(\mathbf{X},\mathbf{x})
\end{equation}

where $\mathbf{k}(\mathbf{x},\mathbf{X})$ is a $n$-dimensional vector such that its $i$th component is $k\big(\mathbf{x},\mathbf{x}^{(i)}\big)$, $\mathbf{k}(\mathbf{X},\mathbf{x})$ is the transposed vector, and $\mathbf{K}$ is the $n\times n$ kernel matrix with entries $K_{i,j}=k\big(\mathbf{x}^{(i)},\mathbf{x}^{(j)}\big)$. Finally, $\lambda^2$ is used to deal with noisy objective functions as well as to ensure that the matrix inversion operation can be performed.

Fitting a GP model on the dataset $\mathcal{D}$ means conditioning the predictive mean $\mu(\mathbf{x})$ and the predictive uncertainty $\sigma(\mathbf{x})$ -- i.e., the square root of equation~(\ref{eq:s2}) -- to the available dataset $\mathcal{D}$, that is tuning the kernel's hyperparameters (i.e., $\sigma_f^2$ and $\ell$ in the SE kernel) and in case $\lambda^2$, usually via MLE or MAP.
Equations~(\ref{eq:mu}) and~(\ref{eq:s2}) should be indeed intended as $\mu(\mathbf{x})^{(n)}$ and $\sigma^2(\mathbf{x})^{(n)}$, but we omit the suffix to keep the notation as simple as possible.\\

The resulting GP is the model generically denoted with $\mathcal{M}$ in \textbf{Algorithm~\ref{algo:bo}}, which the acquisition function $\mathcal{A_M}(\mathbf{x})$ is based on. According to the basic  BO algorithm, the next solution to evaluate is:
\begin{equation}
    \label{eq:acqfun}
    \mathbf{x}^{(n+1)} = \underset{\mathbf{x}\in\Omega}{\text{optimize }} \mathcal{A_M}(\mathbf{x})
\end{equation}

Although most of the acquisition functions are expressed in terms of \textit{utility} and are, consequently, maximized, some of 
them are expressed in terms of \textit{optimistic} estimation of the objective function; indeed, they are maximized or minimized according to the original optimization problem. For this reason we use the general notation ``optimize'' in equation~(\ref{eq:acqfun}).

\subsection{Improvement-based Acquisition Functions}
Instead of following the chronological order, we start by introducing the GP-based Confidence Bound (GP-CB) acquisition function. It is also known as \textit{optimistic policy}, because it provides the most optimistic approximation of the objective function, depending on $\mathcal{M}$. More specifically, the Lower Confidence Bound (GP-LCB) is used for minimization problems and it is defined as:
\begin{equation}
    \label{eq:lcb}
    LCB(\mathbf{x}) = \mu(\mathbf{x}) - \sqrt \beta \; \sigma(\mathbf{x})
\end{equation}
In the case of a maximization problem, the Upper Confidence Bound (GP-UCB) is obtained by simply replacing the difference with the sum. The role of the uncertainty bonus -- represented by the predictive uncertainty $\sigma(\mathbf{x})$ -- is clear, as well as that of the hyperparameter $\beta$: the GP-LCB addresses the exploitation-exploration trade-off as a scalarization of the two objectives (i.e., exploitation and exploration).

Most of the research focused on how to chose a suitable value for $\beta$ over BO iterations. Here we summarize the most relevant methods (against which we have compared our approach).

\begin{itemize}
    \item \textbf{\cite{srinivas2012information} Theorem 1}. The Search Space $\Omega$ consists of a finite number of choices. It is always possible to consider this case by explicitly taking into account the numerical precision used to represent solutions. The proposed scheduling for $\beta$ is given by:
    \begin{equation}
        \beta^{(n)} = 2 \log\left(\vert G \vert n^2\pi^2 / (6\delta) \right)
    \end{equation}
    with $\vert G \vert$ the number of finite solutions in $\Omega$, and with $\delta \in (0,1)$ such that the probability that the \textit{regret} is sub-linear is not lower than $1-\delta$. As well-known, regret is one of the metric used to analyse the convergence of global optimization algorithms. It is briefly recalled in the following Section~\ref{sec:2.3}.
    \item \textbf{\cite{srinivas2012information} Theorem 2.} The Search Space $\Omega$ is continuous, with an infinite number of solutions. In this case, the proposed scheduling for $\beta$ is:
    \begin{equation}
        \beta^{(n)} = 2 \log\left(2n^2\pi^2/(3\delta)\right) + 2d\left(n^2dbr\sqrt{(\log(4d)/\delta))}\right)
    \end{equation}
    with $r: \Omega=[0,r]^d$, and $a,b>0$ some constants to obtain a sub-linear regret with probability $1-\delta$.
    \item The strongest criticism of the previous scheduling was remarked in \cite{berk2020randomised}: ``\textit{the selection of $\beta$ is not done to optimally balance exploration and exploitation, but is done such that the cumulative regret is bounded. [...] While the regret bound provided [...] is desirable, it unfortunately is far larger than needed. This leads to sub-optimal real world performance due to over-exploration".}\footnote{Indeed, in~\cite{srinivas2012information} the authors empirically divide $\beta$ by 5 just to achieve better results.} On the contrary, the \textbf{Randomised GP-CB} proposed in~\textbf{\cite{berk2020randomised}} samples smaller $\beta$ values from a Gamma distribution whose parameters are set to ensure sub-linear convergence. Specifically:
        \begin{equation}
        \beta^{(n)} \sim \Gamma\big(\kappa^{(n)},\theta\big)
    \end{equation}
    with the shape parameter $\displaystyle \kappa^{(n)}=\frac{\log\left(\frac{n^2+1}{\sqrt{2\pi}}\right)}{\log(1+\theta/2)}$. On the contrary, the scale parameter $\theta$ has to be fine-tuned for each target problem. Basically, increasing $\theta$ will increase exploration.\\
\end{itemize}

Other approaches are not based on the GP-CB and, instead, propose to switch -- randomly or according to a prefixed schema -- between different selection mechanisms. The approach proposed in~\cite{de2021greed} was inspired by the well-known $\varepsilon$-greedy strategy adopted in Reinforcement Learning. Specifically, two alternative methods have been proposed:
\begin{itemize}
    \item \textbf{$\varepsilon$-PF selection}. In this case the next decision $\mathbf{x}^{(n+1)}$ is the optimizer of $\mu(\mathbf{x})$, with probability $1-\varepsilon$, or randomly selected from the Pareto Front (PF) defined over the two objectives $\mu(\mathbf{x})$ and $\sigma(\mathbf{x})$, with probability $\varepsilon$. Recalling results from~\cite{vzilinskas2019bi}, this means using the GP-CB with a randomly selected value for $\beta$.    
    \item \textbf{$\varepsilon$-RS selection.} In this case the next decision $\mathbf{x}^{(n+1)}$ is the optimizer of $\mu(\mathbf{x})$, with probability $1-\varepsilon$, or uniformly selected at random within the entire Search Space $\Omega$, with probability $\varepsilon$.
\end{itemize}

Analogously,~\cite{benjamins2022pi} have recently proposed a selection mechanism which does not rely on the GP-CB but, contrary to~\cite{de2021greed}, their algorithm switches between two well-known acquisition functions, specifically $PI$ and $EI$:
\begin{equation}
    \label{eq:PI}
    PI(\mathbf{x}) = \Phi\left(\frac{\Delta(\mathbf{x})}{\sigma(\mathbf{x})}\right)
\end{equation}
\begin{equation}
    \label{eq:EI}
    EI(\mathbf{x}) =
    \begin{cases}
        \;\Delta(\mathbf{x})\; \Phi\left(\frac{\Delta(\mathbf{x})}{\sigma(\mathbf{x})}\right) + \sigma(\mathbf{x})\; \phi\left(\frac{\Delta(\mathbf{x})}{\sigma(\mathbf{x})}\right) & \text{ if } \sigma(\mathbf{x})>0 \\
        \;0 & \text{ if } \sigma(\mathbf{x})=0
    \end{cases}    
\end{equation}

where $\Delta(\mathbf{x}) = \underset{i=1,\dots,n}{\min}\big\{y^{(i)}\big\}-\mu(\mathbf{x})$ if the target problem is a minimization problem (otherwise $\Delta(\mathbf{x}) = \mu(\mathbf{x})-\underset{i=1,\dots,n}{\max}\big\{y^{(i)}\big\}$), and $\Phi(\mathbf{x})$ and $\phi(\mathbf{x})$ are the standard Gaussian cumulative and probability density function, respectively.
We have reported here the two formulations adopted by the authors in~\cite{benjamins2022pi}. However, it is important to remark that some modifications have been proposed in literature to increase exploration, especially for $PI$.

Moreover, two mechanisms have been proposed and investigated by the authors: \textit{alternating} (aka \textit{round-robin}) between the two acquisition functions and starting with $EI$ to finally switch to $PI$ when a certain percentage of the overall evaluations is reached. It is important to remark that the last scheduling is completely opposite to (Srinivas): it is more explorative at the beginning and more exploitative at the end, while the logic underlying the scheduling proposed in~\cite{srinivas2012information}  increases the chance for exploration with the BO iterations.

\subsection{The disregarded relevance of the discrepancy}
\label{sec:2.3}
\textbf{Regret as convergence metric.} As previously mentioned, regret is typically used as a natural performance metric for global optimization algorithms. The \textit{instantaneous regret} at iteration $n$, associated to the decision $\mathbf{x}^{(n)}$, is $r^{(n)} = y^*-y^{(n)}$, with $y^* = f(\mathbf{x}^*)+\varepsilon$. The \textit{cumulative regret}, up to iteration $N$, is $R_N=\sum_{n=1}^N r^{(n)}$. A desirable property is $\lim_{N\rightarrow\infty}R_N=0$ (i.e., no-regret). For finite $N$, the value $R_N/N$ translates into convergence rate. The critical issue is that over-exploration (as implied by the scheduling of $\beta$ in~\cite{srinivas2012information}) increases the value $R_N/N$ to avoid to get stuck into local optima, while other methods decrease it, but at the cost of a higher chance to achieve a sub-optimal solution.

Moreover, in a global optimization task, minimizing the cumulative regret can be counter-intuitive from a human search point of view. If $y^*$ is achieved after few iterations, a human searcher will not continue to evaluate solutions close to the current best one (even if he/she does not known the value of $y^*$, a-priori). Instead, she/he will spend some evaluations to reduce uncertainty over the search space~\cite{candelieri2021uncertainty,candelieri2022explaining}. This because the task she/he is solving is \textit{searching for the best} instead of accumulating some reward. It is the \textit{feeling} about the impossibility to improve that triggers pure exploration in a human searcher, but this behaviour is not modelled by traditional acquisition function~\cite{candelieri2021uncertainty,candelieri2022explaining}.\\

\textit{At the very end, an acquisition function is effective if it provides a balanced trade-off between exploration and exploitation, \textbf{over the entire optimization process}}.\\

\begin{figure}[h!]
    \centering
    \includegraphics[width=\textwidth]{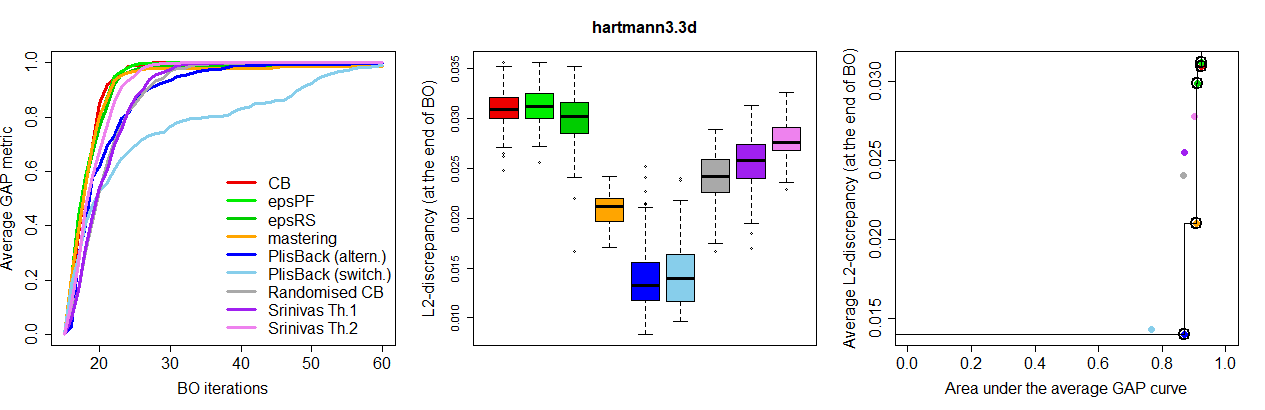}
    \caption{Comparing BO based on different improvement based acquisition functions: (left) GAP metric curves averaged on 100 independent runs, (middle) L2-discrepancy at the end of the optimization process and over 100 independent runs, (right) Pareto analysis between area under the average GAP metric curve and L2-discrepancy.}
    \label{fig:1}
\end{figure}

\noindent
\textbf{GAP metric.} To better explain the previous statement, we first introduce the GAP metric, another common performance indicator to compare optimization algorithms.
\begin{equation}
    GAP_n = \frac{\big\vert y^+_0 - y^{+(n)}\big\vert}{\vert y^+_0 - y^*\vert}
\end{equation}

with $y^+_0=\text{best}\Big\{y^{(1)},...,y^{(n_0)}\Big\}$ and $y^{+(n)}=\text{best}\Big\{y^{(n_0+1)},...,y^{(n)}\Big\}$ (with ``best'' replaced with ``$\max$'' or ``$\min$'' according to the original optimization problem; the absolute value is adopted to independently deal with both maximization and minimization). The GAP metric has important features: it is defined in $[0,1]$ so it can be used to compare different approaches also on completely different optimization problems, and it allows to clearly identify at which iteration the best solution has been identified. However, it is not directly linked to $R_N/N$. Indeed, consider Figure~\ref{fig:1}, that is an anticipation of the results presented in the next. The chart on the left depicts the GAP curves (i.e., $GAP_n$ with $n=n_0,\dots,N$) for BO using different acquisition functions. More precisely, each curve is the average GAP curve on 100 independent runs. Four of them achieve, almost immediately, the GAP value equal to $1$ (i.e., CB, epsPF, epsRS, and mastering), but do not provide any information about the successive decisions and, consequently, the value of $R_N/N$. However, at the end of each optimization process one can observe how many acquisition functions have used the remaining trials to explore -- similarly to a human searcher -- instead of exploiting without any further improvement. To quantify the capacity of exploration, we use \textit{discrepancy}~\cite{fang2018theory}, from Design of Experiments.
\vspace{-0.25cm}

\textbf{Discrepancy} measures how far a given distribution of points deviates from a perfectly uniform one. Many discrepancies have been proposed, we have decided to consider the so-called L2-discrepancy because it
can be expressed analytically even for high dimension. Formally:
\begin{equation}
    D_{L2}(\mathbf{X})=\left[\int_{[0;1]^{2d}}\left(
    \frac{A(\mathbf{X};J_{a,b})}{n} - Vol(J_{a,b})\right)^2 dadb\right]^{1/2}
\end{equation}
where $Vol(J_{a,b})$ denotes the volume of a subset $J_{a,b} \subset[0;1]^{2d}$, $A(\mathbf{X};J_{a,b})$ is the number of points of $\mathbf{X}$ falling into $J_{a,b}$, and finally $a,b\in [0;1]^d$ and $J_{a,b}=[a_1,b_1)\times\dots\times [a_d,b_d)$. With respect to solutions generated by BO, the lower the discrepancy the higher the exploration.

Coming back to Figure~\ref{fig:1}, the charts in the middle shows that only one of the four acquisition functions with the best GAP curves has a low discrepancy, averaged on 100 independent runs (i.e., the so-called ``mastering'').\\

\textbf{Pareto analysis: exploration-exploitation balance.} To finally evaluate the effectiveness of the searching process implied by any acquisition function in terms of both convergence to the optimum and exploration capability, we suggest to perform a Pareto analysis. While it is natural to obtain a single value for the discrepancy (i.e., the average over 100 independent runs), we need to also condensate a GAP curve into a single scalar value. We suggest to compute the area under the GAP curve -- analogously to the AUC, Area Under the ROC Curve used to compare Machine Learning algorithms -- that is:
\begin{equation}
    A_{GAP} = \frac{1}{N-n_0} \sum_{n=n0+1}^N GAP_n
\end{equation}

It is important to remark that the higher the $A_{GAP}$ the better the GAP curve. Coming back again to Figure~\ref{fig:1}, on the right every point represents a different acquisition function, providing a different pair of $A_{GAP}$ (to be maximized) and L2-discrepancy (to be minimized). Only some acquisition functions are Pareto optimal (i.e., circled ones), with just one in the middle of the PF, representing the most well-balanced trade-off between exploration and exploitation.

This analysis will be at the basis of the results presented in this paper. The main benefit is that the effectiveness of a set of acquisition functions can be evaluated in terms of the exploration-exploitation balance over the entire optimization process, instead of the single decision.

\section{An acquisition function mastering the trade-off between exploration and exploitation}
\label{sec:3}
Here we describe our proposed acquisition function, able to implement a BO process well-balanced in terms of exploration and exploitation, overall. Our acquisition function is based on the idea that the most effective and efficient exploitation-exploration trading-off mechanism does not not follow any prefixed or random scheduling: it depends on the evolution of the optimization process, just like in human search. 
Starting from this consideration, we can state that, after a solution is evaluated, only one of the following three possible feedback can be experienced:
\begin{enumerate}
    \item $\mathbf{x}^{(n+1)}$ leads to an $y^{(n+1)}$ \textit{inline} with the expectation $\rightarrow$ \textit{happy, but not surprised}.
    \item $\mathbf{x}^{(n+1)}$ leads to an $y^{(n+1)}$ \textit{better} than  expected $\rightarrow$ \textit{surprisingly happy!}
    \item $\mathbf{x}^{(n+1)}$ leads to an $y^{(n+1)}$ \textit{worse} than  expected $\rightarrow$ \textit{astonished!}\\
\end{enumerate}

Now we try to map these three situations in terms of implied modifications to the model $\mathcal{M}$ and, consequently, the decision suggested by the acquisition function $\mathcal{A_M}(\mathbf{x})$.
\begin{enumerate}
    \item $\mathcal{M}$ will not significantly change, so the next decision will be biased towards exploitation (i.e., local search), if the uncertainty bonus is not large enough.
    \item although we have been collecting a solution better than expected, this means that our model was not so accurate, consequently the updated $\mathcal{M}$ could be significantly different from the previous. As a result, the next decision could be far away from the last evaluated solution, so some exploration (i.e., global search) is triggered by our last decision, independently on the uncertainty bonus.
    \item the only difference with the previous case is that we have been collecting a solution worse than expected, but $\mathcal{M}$ will change anyway, triggering some exploration with respect to the last decision.\\
\end{enumerate}

Summarizing, if $\mathcal{M}$ results inaccurate with respect to the last decision then the next decision could be \textit{explorative}, meaning that it is far from the previous. On the contrary, if the $\mathcal{M}$ is \textit{locally} accurate then the next decision will be biased towards \textit{exploitative} decision, and the uncertainty bonus -- if not sufficiently large -- just ensures to searching within a neighbourhood of the current best.
Indeed, it is important to remark that $\mathcal{M}$ might fail to accurately approximate the objective function in promising regions because observations are sparse in that regions. This is the main motivation underlying the scheduling of $\beta$, in GP-CB, initially proposed by~\cite{srinivas2012information}.\\

\textbf{Algorithm~\ref{algo:master}} summarizes our approach. Basically it works as follows: a pure exploitative decision is obtained by searching for the optimum of the GP's predictive mean, $\mu(\mathbf{x})$. Then the \textit{usefulness} of this decision is considered: if it belongs to the neighbourhood of the current best solution and this neighbourhood already contains a sufficient number of solutions, than the decision is discarded and replaced by a decision aimed at minimizing the overall uncertainty, trying to cover unexplored regions of the search space.

We introduce the following useful notations: $\mathcal{H}_w(\mathbf{x}^+)$ is the neighbourhood of the current best solution $\mathbf{x}^+$, that is a hyper-rectangle centered in $\mathbf{x}^+$ and with sides $w\in\mathbb{R}^d$ (if the search space $\Omega$ is rescaled in $[0;1]^d$, as in the experiments in this paper, then $\mathcal{H}_w(\mathbf{x}^+)$ is a hypercube with side equal to $w\in\mathbb{R}$), $\eta\in\mathbb{N}_+$ is the threshold on the number of solutions belonging to $\mathcal{H}_w(\mathbf{x}^+)$ to discard the pure exploitative decision in favour of a pure explorative one. Although they could be though as hyperparameters of the approach, we provide a reasonable rule-of-thumb to set them avoiding expensive fine-tuning procedures.
Finally, we also introduce the uncertainty quantification to minimize in the case that the exploitative decision is discarded. It is known as Inverse Distance Weighting (IDW), already adopted in global optimization~\cite{bemporad2020global} and empirically resulted more coherent in modelling uncertainty quantification in human search~\cite{candelieri2021uncertainty,candelieri2022explaining}. Moreover, contrary to GP's predictive uncertainty, $\sigma(\mathbf{x})$, IDW is model-free and is able to avoid the variance starvation issue.\color{black}. The IDW, denoted with $z(\mathbf{x})$, is defined as follows.

\begin{equation}
    z(\mathbf{x}) =
    \begin{cases}
        0 & \text{if } \mathbf{x} \in \mathbf{X}\\
        \frac{2}{\pi} \tan^{-1}\left(\frac{1}{\sum_{i=1}^np_i(\mathbf{x})}\right) & \text{ otherwise}
    \end{cases}
\end{equation}

\noindent
where $p_i(\mathbf{x})=\frac{e^{-\|\mathbf{x}-\mathbf{x}^{i}\|^2}}{\|\mathbf{x}-\mathbf{x}^{i}\|^2}$.\\

Another important feature of our algorithm is a \textit{final refining} phase, meaning that the very last decisions are all pure exploitative. This is inline with more recent approaches, such as~\cite{benjamins2022pi}, and contrary to the historical ones, such as~\cite{srinivas2012information}. Indeed, our algorithm does not need to increase exploration with the BO iterations, because it is automatically triggered in the case of a convergence to local optima. However, in the case of an objective function excessively triggering exploration, there could be helpful to spend the last BO iterations in improving (aka refining) the best solution observed so far.

\RestyleAlgo{ruled}

\begin{algorithm}
\caption{Mastering exploration-exploitation trade-off in BO}\label{algo:master}
\KwData{\\ \quad$\Omega=[0;1]^d$ the Search Space (rescale the original search space in case);
\\ \quad$\mathcal{D}=\big\{\big(\mathbf{x}^{(i)},y^{(i)}\big)\big\}$, an initial set of $n$ observations from LHS, and $y^{(i)}=f\big(\mathbf{x}^{(i)}\big)+\varepsilon^{(i)}$,\\\quad\quad with $\varepsilon^{(i)}\sim \mathcal{N}(0,\lambda^2)$;
\\ \quad$N_1>N_2: N_1+N_2=N>n$.}
 \While{$n<N_1$}{
     $(\mathbf{x}^+,y^+): y^+=\text{best}\big\{y^{(i)}:\big(\mathbf{x}^{(i)},y^{(i)}\big) \in \mathcal{D}\big\}$\;
     $\mathcal{H}_w(\mathbf{x}^+)$ is an hypercube centered in $\mathbf{x}^+$ with side equal to $w$\;
     fit the GP model on $\mathcal{D}$ and obtain $\mu(\mathbf{x})$ and $\sigma(\mathbf{x})$\;
     $\mathbf{x}^{(n+1)} = \underset{\mathbf{x}\in\Omega} {\text{optimize }} \mu(\mathbf{x}) \quad\quad\quad\quad\vartriangleleft$ exploitative decision\;
     \If{$\mathbf{x}^{(n+1)}\in \mathcal{H}_w(\mathbf{x}^+) \wedge \vert\mathcal{H}_w(\mathbf{x}^+)\vert \geq \eta$} {$\mathbf{x}^{(n+1)} = \underset{\mathbf{x}\in\Omega} {\text{optimize }} z(\mathbf{x}) \quad\quad\vartriangleleft$ explorative decision}
     observe $y^{(n+1)}=f\big(\mathbf{x}^{(i)}\big)+\varepsilon^{(i)}$\;
     $\mathcal{D} \leftarrow \mathcal{D} \cup \big\{\big(\mathbf{x}^{(n+1)},y^{(n+1)}\big)\big\}$\;
     $n \leftarrow n+1$
 }
 \CommentSty{*** final refining: only exploitation ***}\\
 \While{$n<N_1+N_2$} {
     fit the GP model on $\mathcal{D}$ and obtain $\mu(\mathbf{x})$ and $\sigma(\mathbf{x})$\;
     $\mathbf{x}^{(n+1)} = \underset{\mathbf{x}\in\Omega} {\text{optimize }} \mu(\mathbf{x})$\;
     observe $y^{(n+1)}=f\big(\mathbf{x}^{(i)}\big)+\varepsilon^{(i)}$\;
     $\mathcal{D} \leftarrow \mathcal{D} \cup \big\{\big(\mathbf{x}^{(n+1)},y^{(n+1)}\big)\big\}$\;
     $n \leftarrow n+1$
 }
 \KwResult{\\ \quad$(\mathbf{x}^+,y^+): y^+=\text{best}\big\{y^{(i)}:\big(\mathbf{x}^{(i)},y^{(i)}\big) \in \mathcal{D}\big\}$}  
\end{algorithm}

\section{Experimental Setting}
Experiments consider 10 well-known global optimization test functions, widely used in literature, and with search spaces of different dimensionalities: Branin ($d=2$), Camel3 ($d=3$), Camel6 ($d=6$), GoldPr ($d=2$), Hartmann3 ($d=3$), Hartmann4 ($d=4$), Hartmann6 ($d=6$), Rosenbrock ($d=2$), Schwefel ($d=2$), StybTang ($d=2$).

The proposed approach has been compared against 8 state-of-the-art improvement-based acquisition functions, specifically: CB with constant $\beta$ (i.e., $\beta=1$), CB with $\beta$ scheduled as in Theorem 1 and Theorem 2 of~\cite{srinivas2012information}, Randomised GP-CB~\cite{berk2020randomised}, $\varepsilon$RS and $\varepsilon$PF from~\cite{de2021greed}, alternating and switching between EI and PI as proposed in~\cite{benjamins2022pi}.

As far as the hyperparameters of the considered BO algorithms are concerned, every paper shows that a unique optimal configuration does not exist. For each algorithm, the hyperparameters configuration offering the best performances on the largest number of test problems has been selected\footnote{All the hyperparameters configurations are explicitly reported in the code, freely available, as detailed in Section~\ref{sec:6}}. For the proposed algorithm, the suggested hyperparameters values are $w=0.1$ and $\eta=\lfloor15\times d\rfloor/3$.

For each test problem, and for each BO algorithm, 100 independent runs have been performed. In each run the algorithms share the same initialization of $\mathcal{D}$. The initial size of $\mathcal{D}$ is $5 \times d$, while the overall number of function evaluations is $N=20\times d$. For the proposed algorithm, we split into $N_1=15\times d$ and $N_2=5\times d$ (for the refining phase).

As initially mentioned, the probabilistic model is a GP regression model with SE kernel. All the acquisition functions are optimized following a quite common approach: sampling $100\times d$ solutions via LHS and then optimizing the $5$ most promising ones through L-BFGS-B.

All the experiments have performed on a Intel(R) Core(TM) i7-7700HQ, 2.80GHz, 16.0GB, Windows 10 (64bit). The code is implemented in R (version 4.2.3, 2023-03-15 ucrt -- "Shortstop Beagle") and run in RStudio (2023.03.0 Build 386). All the other details are reported in Section~\ref{sec:6}.

\section{Results}
This section summarizes the empirical results obtained from the experiments. We started with the most important result, concisely summarized in \textbf{Table 1}: a cross indicates that a certain algorithm (column) resulted Pareto optimal on a certain test problem (row).

Surprisingly, CB with constant $\beta=1$ resulted Pareto optimal more times than other CB-based acquisition functions. However, it is important to remark that Pareto optimality here refers to the exploration-exploitation trade-off: an acquisition function leading to sub-optimal solutions (i.e., low area under the average GAP metric) could anyway result Pareto optimal if it provides really low L2-discrepancy. This happens on the problems Camel3, Camel6, and GoldPr, in which the constant $\beta=1$ is surely higher than those provided by the other CB-based acquisition functions. As we said, high values of $\beta$ lead to over-exploration. On the contrary, on the test problems Hartmann3, Hartmann4, Hartmann6, and Schwefel, constant $\beta=1$ leads to under-exploration when compared to~\cite{srinivas2012information} Theorem 1 and Theorem 2.

\begin{figure}[h!]
    \centering
    \includegraphics[width=\textwidth]{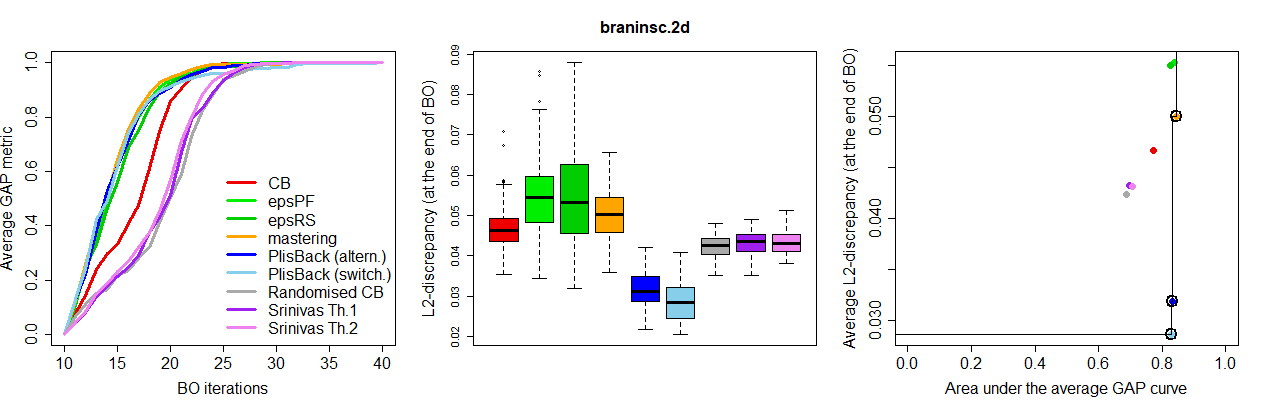}
    \caption{Branin test function: average GAP metric curves (on the left), L2-discrepancy at the end of the optimization processes (in the middle), and Pareto analysis between area under the GAP curve and average L2-discrepancy (on the right).}
    \label{fig:2}
\end{figure}

\begin{figure}[h!]
    \centering
    \includegraphics[width=\textwidth]{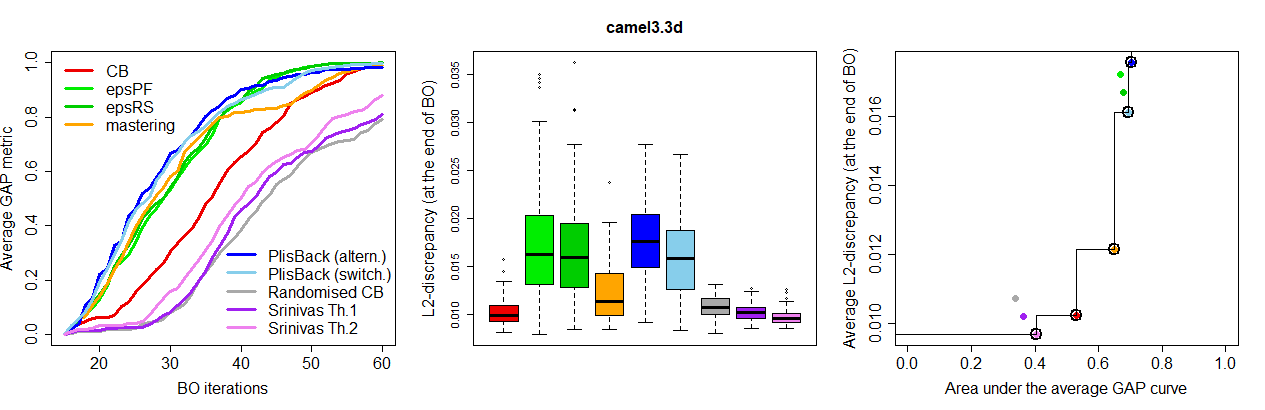}
    \caption{Camel3 test function: average GAP metric curves (on the left), L2-discrepancy at the end of the optimization processes (in the middle), and Pareto analysis between area under the GAP curve and average L2-discrepancy (on the right).discrepancy.}
    \label{fig:3}
\end{figure}

\begin{figure}[h!]
    \centering
    \includegraphics[width=\textwidth]{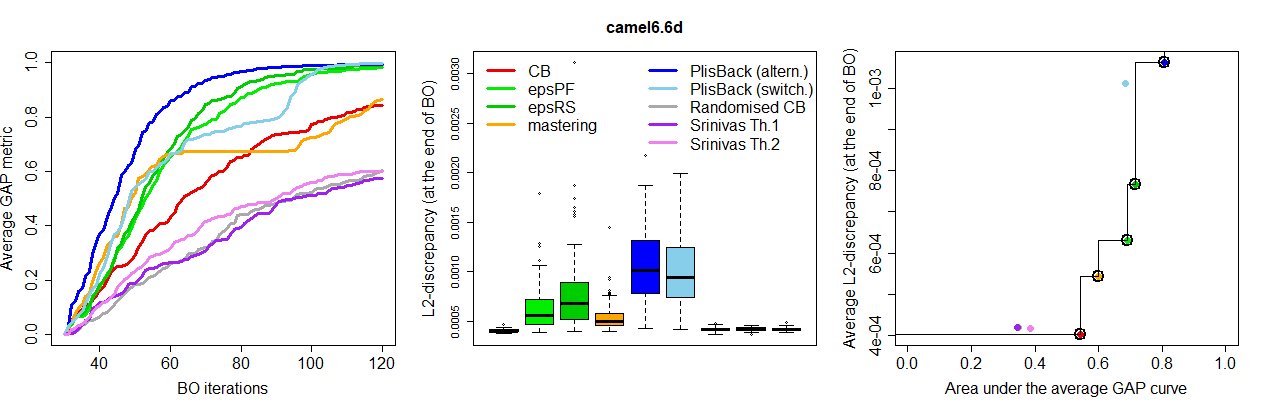}
    \caption{Camel6 test function: average GAP metric curves (on the left), L2-discrepancy at the end of the optimization processes (in the middle), and Pareto analysis between area under the GAP curve and average L2-discrepancy (on the right).}
    \label{fig:4}
\end{figure}

\begin{figure}[h!]
    \centering
    \includegraphics[width=\textwidth]{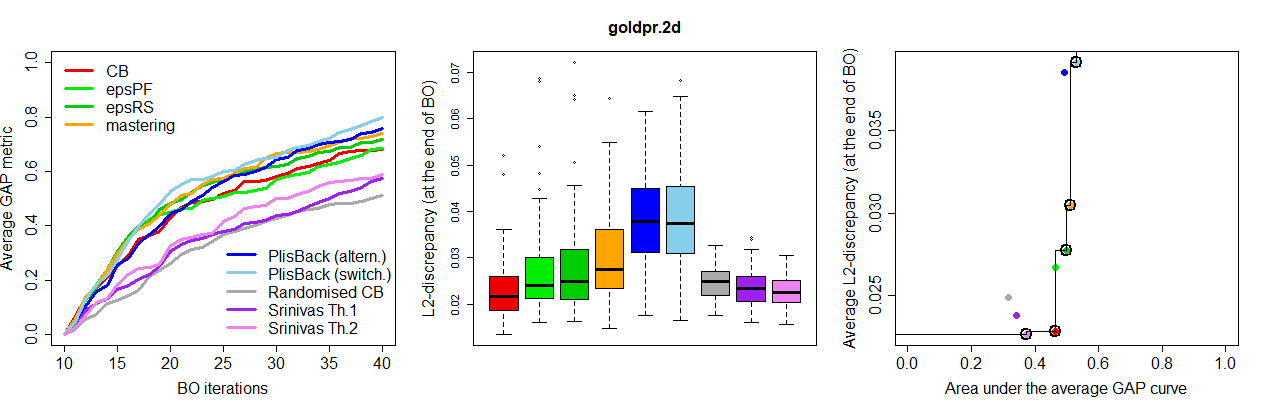}
    \caption{GoldPr test function: average GAP metric curves (on the left), L2-discrepancy at the end of the optimization processes (in the middle), and Pareto analysis between area under the GAP curve and average L2-discrepancy (on the right).}
    \label{fig:5}
\end{figure}

\begin{figure}[h!]
    \centering
    \includegraphics[width=\textwidth]{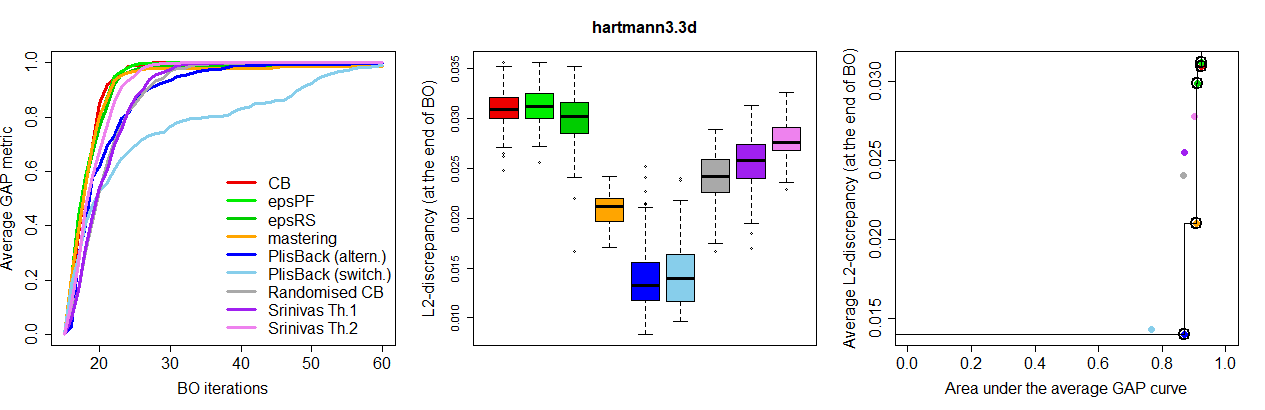}
    \caption{Hartmann3 test function: average GAP metric curves (on the left), L2-discrepancy at the end of the optimization processes (in the middle), and Pareto analysis between area under the GAP curve and average L2-discrepancy (on the right).}
    \label{fig:6}
\end{figure}

\begin{figure}[h!]
    \centering
    \includegraphics[width=\textwidth]{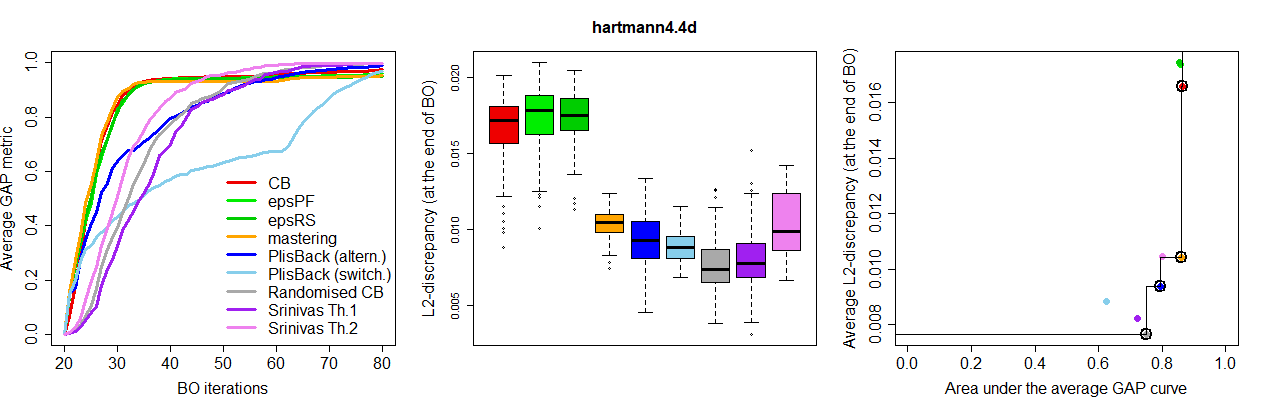}
    \caption{Hartmann4 test function: average GAP metric curves (on the left), L2-discrepancy at the end of the optimization processes (in the middle), and Pareto analysis between area under the GAP curve and average L2-discrepancy (on the right).}
    \label{fig:7}
\end{figure}

\begin{figure}[h!]
    \centering
    \includegraphics[width=\textwidth]{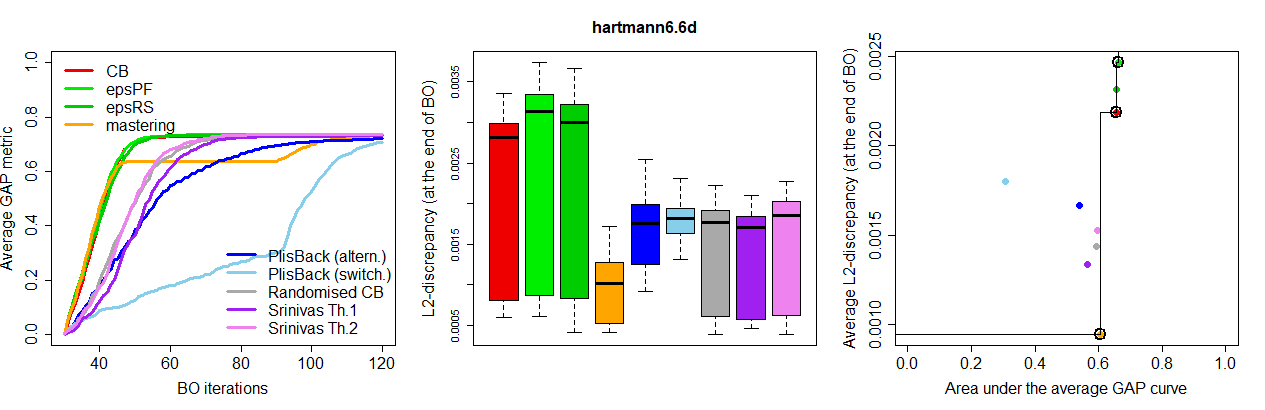}
    \caption{Hartmann6 test function: average GAP metric curves (on the left), L2-discrepancy at the end of the optimization processes (in the middle), and Pareto analysis between area under the GAP curve and average L2-discrepancy (on the right).}
    \label{fig:8}
\end{figure}

\begin{figure}[h!]
    \centering
    \includegraphics[width=\textwidth]{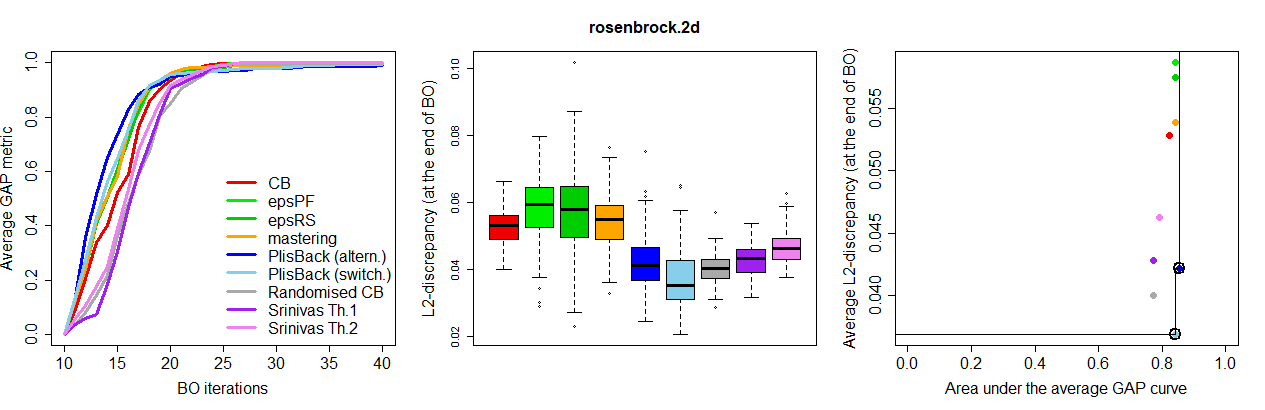}
    \caption{Rosenbrock test function: average GAP metric curves (on the left), L2-discrepancy at the end of the optimization processes (in the middle), and Pareto analysis between area under the GAP curve and average L2-discrepancy (on the right).}
    \label{fig:9}
\end{figure}

\begin{figure}[h!]
    \centering
    \includegraphics[width=\textwidth]{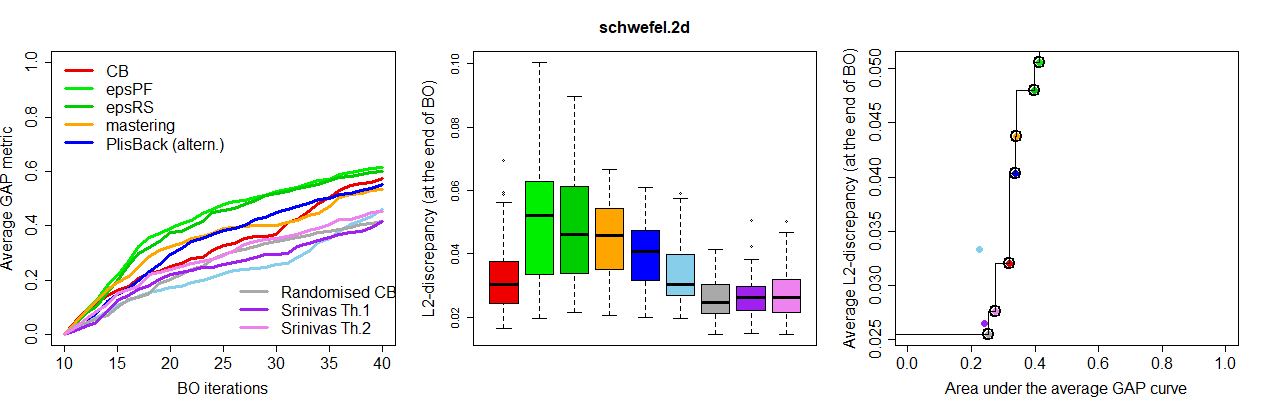}
    \caption{Schwefel test function: average GAP metric curves (on the left), L2-discrepancy at the end of the optimization processes (in the middle), and Pareto analysis between area under the GAP curve and average L2-discrepancy (on the right).}
    \label{fig:10}
\end{figure}

\begin{figure}[h!]
    \centering
    \includegraphics[width=\textwidth]{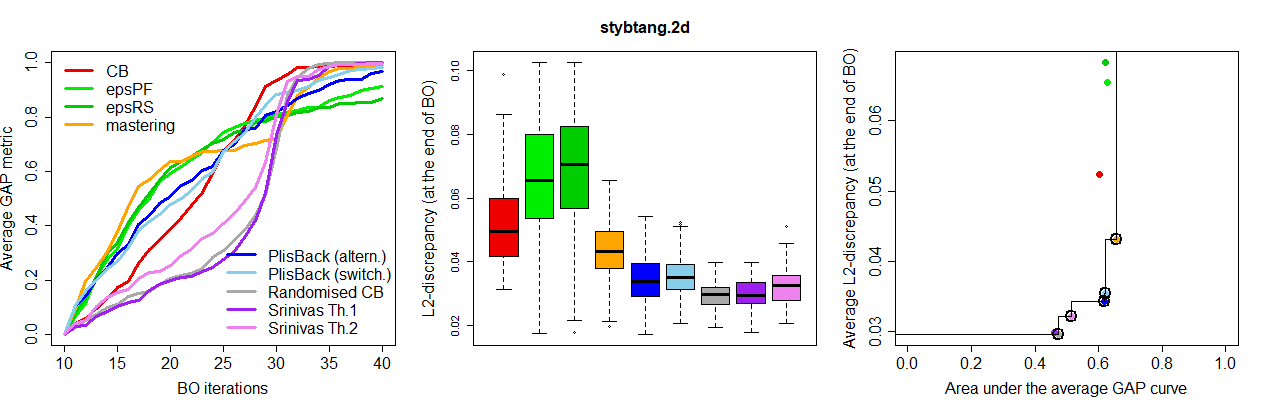}
    \caption{StybTang test function: average GAP metric curves (on the left), L2-discrepancy at the end of the optimization processes (in the middle), and Pareto analysis between area under the GAP curve and average L2-discrepancy (on the right).}
    \label{fig:11}
\end{figure}

We can conclude that, although Pareto optimal on 7 out of 10 test problems, CB with constant $\beta=1$ was almost always located at the extreme of the Pareto Front, consequently representing a not well-balanced trade-off between exploration and exploitation. This can be observed in the Figures~\ref{fig:2} to~\ref{fig:11}. Every figure consists of three charts:
\begin{itemize}
    \item \textbf{on the left:} the GAP curve averaged on 100 independent runs is depicted for each algorithm. Standard deviations are omitted for a clear visualization -- in appendix, the box plots of the GAP value at the end of the optimization processes are reported: there are not significant differences among methods. The first point of the curves refers to the best solution over the LHS-initialized $\mathcal{D}$.
    \item \textbf{in the middle:} the box-plot of the L2-discrepancies obtained over the 100 independent runs, separately for each algorithm.
    \item \textbf{on the right:} Pareto analysis among the algorithms with respect to: (\textit{a}) area under the average GAP curve (converge to the optimum) and (\textit{b}) L2-discrepancy (capacity to explore).
\end{itemize}

\begin{table}[h!]
\caption{Pareto optimal acquisition functions on 10 global optimization test problems. A cross is present if the acquisition function resulted Pareto optimal on the problem.}
\resizebox{\columnwidth}{!}{%
\begin{tabular}{lccccccccc}
\hline
\multicolumn{1}{c}{}
& \textbf{CB} & \textbf{Srinivas et al.} & \textbf{Srinivas et al.} & \textbf{Randomised} & \textbf{$\varepsilon$RS} & \textbf{$\varepsilon$PF} & \textbf{PI is Back!} & \textbf{PI is back!}  & \textbf{Mastering} \\
& \textbf{const. $\beta$} & \textbf{Theorem 1} & \textbf{Theorem 2} & \textbf{CB} & \textbf{} & \textbf{} & \textbf{alternating} & \textbf{switching} & \textbf{(proposed)}\\ 
\hline
\multicolumn{1}{l|}{\textbf{Branin ($d=2$)}} & & & & & & & x & x & x\\
\multicolumn{1}{l|}{\textbf{Camel3 ($d=3$)}} & x & & x & & x & & x & & x\\
\multicolumn{1}{l|}{\textbf{Camel6 ($d=6$)}} & x & & & & x & x & x & & x \\
\multicolumn{1}{l|}{\textbf{GoldPr ($d=2$)}} & x & & x & & x & & & x & x \\
\multicolumn{1}{l|}{\textbf{Hartmann3 ($d=3$)}} & x & & & & x & x & x & & x \\
\multicolumn{1}{l|}{\textbf{Hartmann4 ($d=4$)}} & x & & & x & & & x & & x \\
\multicolumn{1}{l|}{\textbf{Hartmann6 ($d=6$)}} & x & & & & & x & & & x \\
\multicolumn{1}{l|}{\textbf{Rosenbrock ($d=2$)}} & & & & & & & x & x & \\
\multicolumn{1}{l|}{\textbf{Schwefel ($d=2$)}} & x & & x & x & x & x & x & & x \\
\multicolumn{1}{l|}{\textbf{StybTang ($d=2$)}} & & & x & x & & & x & x & x \\
\hline
& 7/10 & 0/10 & 4/10 & 3/10 & 5/10 & 4/10 & 8/10 & 4/10 & 9/10 \\
\hline
\end{tabular}%
}
\end{table}

Another acquisition function resulted Pareto optimal on a large number of test problems is the alternating schema between $EI$ and $PI$~\cite{benjamins2022pi}. It is interesting to highlight that this acquisition function locates on the opposite side of the Pareto Front with respect to CB with constant $\beta=1$, on all the test problems on which both the algorithms resulted Pareto optimal.\\

Finally, the proposed approach resulted Pareto optimal on all the test problems but one (i.e., Rosenbrock). More important, it is almost always ($6/9$) located in the central part of the Pareto Front (i.e., Camel3, camel6, GoldPr, Hartmann3, Hartmann4, Schwefel). This point is crucial, because it means that the proposed approach -- further than Pareto optimal -- offers the best exploration-exploitation balance.\\

As a final remark, only the method recently proposed in~\cite{benjamins2022pi} (both alternating and switching) resulted Pareto optimal on the Rosenbrock test problem.

\color{black}

\newpage

\section{Replicability of experiments: open-source code}
\label{sec:6}
To guarantee replicability of the experiments, the entire code is available for free at the following link, along with results and figures reported in this paper:

\url{https://github.com/acandelieri/MasteringExplorationExploitationInBO.git}\\

\noindent
Within the code all the values of the hyperparameters for all the methods are also specified.

\section{Conclusions}
Exploration-exploitation balance is still an open challenge in all the learning-and-optimization frameworks, from evolutionary algorithms to Reinforcement Learning and global optimization of black-box expensive function. In this paper the improvement-based acquisition functions adopted in Bayesian Optimization have been investigated, considering both foundational and more recent methods. Limitations have been identified, specifically the fact that all the methods are based on prefixed or random scheduling of the uncertainty bonus to guarantee convergence to the global optimum.

Results and insights from studies on human search suggested how to implement a novel mechanism to dynamically and adaptively switch between exploration and exploitation over the entire optimization process, depending on the feedback collected.

The comparison among the considered acquisition functions has been performed by proposing a Pareto analysis between the convergence to the optimum (exploitation), measured as the area under the GAP metric curve, and the exploration capabilities, over the entire optimization process. 

As a result, the proposed approach resulted Pareto optimal in 9 out of 10 test problems and, even more important, it is usually located in the central part of the Pareto Front, meaning that it is the one providing the most balanced trade-off between exploration and exploitation.

The main limitation of the proposed approach is that, at each iteration, first a pure exploitative decision is suggested (i.e., by optimizing the GP's predictive mean) but, if discarded because resulting useless, another pure explorative decision must be computed (i.e., by minimizing the uncertainty quantified via Inverse Distance Weighting). In the worst case, two internal optimization problems are solved in each BO iteration. 

Although it cannot be considered the \textit{ultimate} acquisition function, the proposed method provides empirical evidence on how the design of adaptive scheduling should be the research direction for the next years, with the aim to achieve the master acquisition function for BO.


\bibliographystyle{ACM-Reference-Format}
\bibliography{master_acquisition}


\begin{thebibliography}{50}


\ifx \showCODEN    \undefined \def \showCODEN     #1{\unskip}     \fi
\ifx \showDOI      \undefined \def \showDOI       #1{#1}\fi
\ifx \showISBNx    \undefined \def \showISBNx     #1{\unskip}     \fi
\ifx \showISBNxiii \undefined \def \showISBNxiii  #1{\unskip}     \fi
\ifx \showISSN     \undefined \def \showISSN      #1{\unskip}     \fi
\ifx \showLCCN     \undefined \def \showLCCN      #1{\unskip}     \fi
\ifx \shownote     \undefined \def \shownote      #1{#1}          \fi
\ifx \showarticletitle \undefined \def \showarticletitle #1{#1}   \fi
\ifx \showURL      \undefined \def \showURL       {\relax}        \fi
\providecommand\bibfield[2]{#2}
\providecommand\bibinfo[2]{#2}
\providecommand\natexlab[1]{#1}
\providecommand\showeprint[2][]{arXiv:#2}

\bibitem[Archetti and Candelieri(2019)]%
        {archetti2019bayesian}
\bibfield{author}{\bibinfo{person}{Francesco Archetti} {and}
  \bibinfo{person}{Antonio Candelieri}.} \bibinfo{year}{2019}\natexlab{}.
\newblock \bibinfo{booktitle}{\emph{Bayesian optimization and data science}}.
  Vol.~\bibinfo{volume}{849}.
\newblock \bibinfo{publisher}{Springer}.
\newblock


\bibitem[Baheri et~al\mbox{.}(2017)]%
        {baheri2017real}
\bibfield{author}{\bibinfo{person}{Ali Baheri}, \bibinfo{person}{Shamir
  Bin-Karim}, \bibinfo{person}{Alireza Bafandeh}, {and}
  \bibinfo{person}{Christopher Vermillion}.} \bibinfo{year}{2017}\natexlab{}.
\newblock \showarticletitle{Real-time control using Bayesian optimization: A
  case study in airborne wind energy systems}.
\newblock \bibinfo{journal}{\emph{Control Engineering Practice}}
  \bibinfo{volume}{69} (\bibinfo{year}{2017}), \bibinfo{pages}{131--140}.
\newblock


\bibitem[Balandat et~al\mbox{.}(2020)]%
        {balandat2020botorch}
\bibfield{author}{\bibinfo{person}{Maximilian Balandat}, \bibinfo{person}{Brian
  Karrer}, \bibinfo{person}{Daniel Jiang}, \bibinfo{person}{Samuel Daulton},
  \bibinfo{person}{Ben Letham}, \bibinfo{person}{Andrew~G Wilson}, {and}
  \bibinfo{person}{Eytan Bakshy}.} \bibinfo{year}{2020}\natexlab{}.
\newblock \showarticletitle{BoTorch: A framework for efficient Monte-Carlo
  Bayesian optimization}.
\newblock \bibinfo{journal}{\emph{Advances in neural information processing
  systems}}  \bibinfo{volume}{33} (\bibinfo{year}{2020}),
  \bibinfo{pages}{21524--21538}.
\newblock


\bibitem[Bemporad(2020)]%
        {bemporad2020global}
\bibfield{author}{\bibinfo{person}{Alberto Bemporad}.}
  \bibinfo{year}{2020}\natexlab{}.
\newblock \showarticletitle{Global optimization via inverse distance weighting
  and radial basis functions}.
\newblock \bibinfo{journal}{\emph{Computational Optimization and Applications}}
  \bibinfo{volume}{77}, \bibinfo{number}{2} (\bibinfo{year}{2020}),
  \bibinfo{pages}{571--595}.
\newblock


\bibitem[Benjamins et~al\mbox{.}(2022)]%
        {benjamins2022pi}
\bibfield{author}{\bibinfo{person}{Carolin Benjamins}, \bibinfo{person}{Elena
  Raponi}, \bibinfo{person}{Anja Jankovic}, \bibinfo{person}{Koen van~der
  Blom}, \bibinfo{person}{Maria~Laura Santoni}, \bibinfo{person}{Marius
  Lindauer}, {and} \bibinfo{person}{Carola Doerr}.}
  \bibinfo{year}{2022}\natexlab{}.
\newblock \showarticletitle{PI is back! Switching Acquisition Functions in
  Bayesian Optimization}.
\newblock \bibinfo{journal}{\emph{arXiv preprint arXiv:2211.01455}}
  (\bibinfo{year}{2022}).
\newblock


\bibitem[Berk et~al\mbox{.}(2020)]%
        {berk2020randomised}
\bibfield{author}{\bibinfo{person}{Julian Berk}, \bibinfo{person}{Sunil Gupta},
  \bibinfo{person}{Santu Rana}, {and} \bibinfo{person}{Svetha Venkatesh}.}
  \bibinfo{year}{2020}\natexlab{}.
\newblock \showarticletitle{Randomised gaussian process upper confidence bound
  for bayesian optimisation}.
\newblock \bibinfo{journal}{\emph{arXiv preprint arXiv:2006.04296}}
  (\bibinfo{year}{2020}).
\newblock


\bibitem[Berk et~al\mbox{.}(2019)]%
        {berk2019exploration}
\bibfield{author}{\bibinfo{person}{Julian Berk}, \bibinfo{person}{Vu Nguyen},
  \bibinfo{person}{Sunil Gupta}, \bibinfo{person}{Santu Rana}, {and}
  \bibinfo{person}{Svetha Venkatesh}.} \bibinfo{year}{2019}\natexlab{}.
\newblock \showarticletitle{Exploration enhanced expected improvement for
  Bayesian optimization}. In \bibinfo{booktitle}{\emph{Machine Learning and
  Knowledge Discovery in Databases: European Conference, ECML PKDD 2018,
  Dublin, Ireland, September 10--14, 2018, Proceedings, Part II 18}}. Springer,
  \bibinfo{pages}{621--637}.
\newblock


\bibitem[Bischl et~al\mbox{.}(2017)]%
        {bischl2017mlrmbo}
\bibfield{author}{\bibinfo{person}{Bernd Bischl}, \bibinfo{person}{Jakob
  Richter}, \bibinfo{person}{Jakob Bossek}, \bibinfo{person}{Daniel Horn},
  \bibinfo{person}{Janek Thomas}, {and} \bibinfo{person}{Michel Lang}.}
  \bibinfo{year}{2017}\natexlab{}.
\newblock \showarticletitle{mlrMBO: A modular framework for model-based
  optimization of expensive black-box functions}.
\newblock \bibinfo{journal}{\emph{arXiv preprint arXiv:1703.03373}}
  (\bibinfo{year}{2017}).
\newblock


\bibitem[Candelieri et~al\mbox{.}(2020)]%
        {candelieri2020learning}
\bibfield{author}{\bibinfo{person}{Antonio Candelieri}, \bibinfo{person}{Bruno
  Galuzzi}, \bibinfo{person}{Ilaria Giordani}, {and} \bibinfo{person}{Francesco
  Archetti}.} \bibinfo{year}{2020}\natexlab{}.
\newblock \showarticletitle{Learning Optimal Control of Water Distribution
  Networks Through Sequential Model-Based Optimization}. In
  \bibinfo{booktitle}{\emph{Learning and Intelligent Optimization: 14th
  International Conference, LION 14, Athens, Greece, May 24--28, 2020, Revised
  Selected Papers 14}}. Springer, \bibinfo{pages}{303--315}.
\newblock


\bibitem[Candelieri et~al\mbox{.}(2018)]%
        {candelieri2018bayesian}
\bibfield{author}{\bibinfo{person}{Antonio Candelieri},
  \bibinfo{person}{Raffaele Perego}, {and} \bibinfo{person}{Francesco
  Archetti}.} \bibinfo{year}{2018}\natexlab{}.
\newblock \showarticletitle{Bayesian optimization of pump operations in water
  distribution systems}.
\newblock \bibinfo{journal}{\emph{Journal of Global Optimization}}
  \bibinfo{volume}{71} (\bibinfo{year}{2018}), \bibinfo{pages}{213--235}.
\newblock


\bibitem[Candelieri et~al\mbox{.}(2021a)]%
        {candelieri2021green}
\bibfield{author}{\bibinfo{person}{Antonio Candelieri},
  \bibinfo{person}{Riccardo Perego}, {and} \bibinfo{person}{Francesco
  Archetti}.} \bibinfo{year}{2021}\natexlab{a}.
\newblock \showarticletitle{Green machine learning via augmented Gaussian
  processes and multi-information source optimization}.
\newblock \bibinfo{journal}{\emph{Soft Computing}} (\bibinfo{year}{2021}),
  \bibinfo{pages}{1--13}.
\newblock


\bibitem[Candelieri et~al\mbox{.}(2021b)]%
        {candelieri2021uncertainty}
\bibfield{author}{\bibinfo{person}{Antonio Candelieri}, \bibinfo{person}{Andrea
  Ponti}, {and} \bibinfo{person}{Francesco Archetti}.}
  \bibinfo{year}{2021}\natexlab{b}.
\newblock \showarticletitle{Uncertainty quantification and
  exploration--exploitation trade-off in humans}.
\newblock \bibinfo{journal}{\emph{Journal of Ambient Intelligence and Humanized
  Computing}} (\bibinfo{year}{2021}), \bibinfo{pages}{1--34}.
\newblock


\bibitem[Candelieri et~al\mbox{.}(2022a)]%
        {candelieri2022explaining}
\bibfield{author}{\bibinfo{person}{Antonio Candelieri}, \bibinfo{person}{Andrea
  Ponti}, {and} \bibinfo{person}{Francesco Archetti}.}
  \bibinfo{year}{2022}\natexlab{a}.
\newblock \showarticletitle{Explaining Exploration--Exploitation in Humans}.
\newblock \bibinfo{journal}{\emph{Big Data and Cognitive Computing}}
  \bibinfo{volume}{6}, \bibinfo{number}{4} (\bibinfo{year}{2022}),
  \bibinfo{pages}{155}.
\newblock


\bibitem[Candelieri et~al\mbox{.}(2022b)]%
        {candelieri2022fair}
\bibfield{author}{\bibinfo{person}{Antonio Candelieri}, \bibinfo{person}{Andrea
  Ponti}, {and} \bibinfo{person}{Francesco Archetti}.}
  \bibinfo{year}{2022}\natexlab{b}.
\newblock \showarticletitle{Fair and green hyperparameter optimization via
  multi-objective and multiple information source bayesian optimization}.
\newblock \bibinfo{journal}{\emph{arXiv preprint arXiv:2205.08835}}
  (\bibinfo{year}{2022}).
\newblock


\bibitem[Candelieri et~al\mbox{.}(2023)]%
        {candelieri2023safe}
\bibfield{author}{\bibinfo{person}{Antonio Candelieri}, \bibinfo{person}{Andrea
  Ponti}, {and} \bibinfo{person}{Francesco Archetti}.}
  \bibinfo{year}{2023}\natexlab{}.
\newblock \showarticletitle{Safe-Exploration of Control Policies from
  Safe-Experience via Gaussian Processes}. In
  \bibinfo{booktitle}{\emph{Learning and Intelligent Optimization: 16th
  International Conference, LION 16, Milos Island, Greece, June 5--10, 2022,
  Revised Selected Papers}}. Springer, \bibinfo{pages}{232--247}.
\newblock


\bibitem[Cuesta~Ramirez et~al\mbox{.}(2022)]%
        {cuesta2022comparison}
\bibfield{author}{\bibinfo{person}{Jhouben Cuesta~Ramirez},
  \bibinfo{person}{Rodolphe Le~Riche}, \bibinfo{person}{Olivier Roustant},
  \bibinfo{person}{Guillaume Perrin}, \bibinfo{person}{C{\'e}dric Durantin},
  {and} \bibinfo{person}{Alain Gli{\`e}re}.} \bibinfo{year}{2022}\natexlab{}.
\newblock \showarticletitle{A comparison of mixed-variables Bayesian
  optimization approaches}.
\newblock \bibinfo{journal}{\emph{Advanced Modeling and Simulation in
  Engineering Sciences}} \bibinfo{volume}{9}, \bibinfo{number}{1}
  (\bibinfo{year}{2022}), \bibinfo{pages}{6}.
\newblock


\bibitem[De~Ath et~al\mbox{.}(2021)]%
        {de2021greed}
\bibfield{author}{\bibinfo{person}{George De~Ath}, \bibinfo{person}{Richard~M
  Everson}, \bibinfo{person}{Alma~AM Rahat}, {and} \bibinfo{person}{Jonathan~E
  Fieldsend}.} \bibinfo{year}{2021}\natexlab{}.
\newblock \showarticletitle{Greed is good: Exploration and exploitation
  trade-offs in Bayesian optimisation}.
\newblock \bibinfo{journal}{\emph{ACM Transactions on Evolutionary Learning and
  Optimization}} \bibinfo{volume}{1}, \bibinfo{number}{1}
  (\bibinfo{year}{2021}), \bibinfo{pages}{1--22}.
\newblock


\bibitem[Dewancker et~al\mbox{.}(2016)]%
        {dewancker2016bayesian}
\bibfield{author}{\bibinfo{person}{Ian Dewancker}, \bibinfo{person}{Michael
  McCourt}, {and} \bibinfo{person}{Scott Clark}.}
  \bibinfo{year}{2016}\natexlab{}.
\newblock \showarticletitle{Bayesian optimization for machine learning: A
  practical guidebook}.
\newblock \bibinfo{journal}{\emph{arXiv preprint arXiv:1612.04858}}
  (\bibinfo{year}{2016}).
\newblock


\bibitem[Diouane et~al\mbox{.}(2022)]%
        {diouane2022trego}
\bibfield{author}{\bibinfo{person}{Youssef Diouane}, \bibinfo{person}{Victor
  Picheny}, \bibinfo{person}{Rodolophe~Le Riche}, {and}
  \bibinfo{person}{Alexandre Scotto~Di Perrotolo}.}
  \bibinfo{year}{2022}\natexlab{}.
\newblock \showarticletitle{TREGO: a trust-region framework for efficient
  global optimization}.
\newblock \bibinfo{journal}{\emph{Journal of Global Optimization}}
  (\bibinfo{year}{2022}), \bibinfo{pages}{1--23}.
\newblock


\bibitem[Eriksson et~al\mbox{.}(2019)]%
        {eriksson2019scalable}
\bibfield{author}{\bibinfo{person}{David Eriksson}, \bibinfo{person}{Michael
  Pearce}, \bibinfo{person}{Jacob Gardner}, \bibinfo{person}{Ryan~D Turner},
  {and} \bibinfo{person}{Matthias Poloczek}.} \bibinfo{year}{2019}\natexlab{}.
\newblock \showarticletitle{Scalable global optimization via local bayesian
  optimization}.
\newblock \bibinfo{journal}{\emph{Advances in neural information processing
  systems}}  \bibinfo{volume}{32} (\bibinfo{year}{2019}).
\newblock


\bibitem[Fang et~al\mbox{.}(2018)]%
        {fang2018theory}
\bibfield{author}{\bibinfo{person}{Kaitai Fang}, \bibinfo{person}{Min-Qian
  Liu}, \bibinfo{person}{Hong Qin}, {and} \bibinfo{person}{Yong-Dao Zhou}.}
  \bibinfo{year}{2018}\natexlab{}.
\newblock \bibinfo{booktitle}{\emph{Theory and application of uniform
  experimental designs}}. Vol.~\bibinfo{volume}{221}.
\newblock \bibinfo{publisher}{Springer}.
\newblock


\bibitem[Frazier et~al\mbox{.}(2009)]%
        {frazier2009knowledge}
\bibfield{author}{\bibinfo{person}{Peter Frazier}, \bibinfo{person}{Warren
  Powell}, {and} \bibinfo{person}{Savas Dayanik}.}
  \bibinfo{year}{2009}\natexlab{}.
\newblock \showarticletitle{The knowledge-gradient policy for correlated normal
  beliefs}.
\newblock \bibinfo{journal}{\emph{INFORMS journal on Computing}}
  \bibinfo{volume}{21}, \bibinfo{number}{4} (\bibinfo{year}{2009}),
  \bibinfo{pages}{599--613}.
\newblock


\bibitem[Frazier(2018)]%
        {frazier2018bayesian}
\bibfield{author}{\bibinfo{person}{Peter~I Frazier}.}
  \bibinfo{year}{2018}\natexlab{}.
\newblock \showarticletitle{Bayesian optimization}.
\newblock In \bibinfo{booktitle}{\emph{Recent advances in optimization and
  modeling of contemporary problems}}. \bibinfo{publisher}{Informs},
  \bibinfo{pages}{255--278}.
\newblock


\bibitem[Garnett(2023)]%
        {garnett2023bayesian}
\bibfield{author}{\bibinfo{person}{Roman Garnett}.}
  \bibinfo{year}{2023}\natexlab{}.
\newblock \bibinfo{booktitle}{\emph{Bayesian optimization}}.
\newblock \bibinfo{publisher}{Cambridge University Press}.
\newblock


\bibitem[Golovin et~al\mbox{.}(2017)]%
        {golovin2017google}
\bibfield{author}{\bibinfo{person}{Daniel Golovin}, \bibinfo{person}{Benjamin
  Solnik}, \bibinfo{person}{Subhodeep Moitra}, \bibinfo{person}{Greg
  Kochanski}, \bibinfo{person}{John Karro}, {and} \bibinfo{person}{David
  Sculley}.} \bibinfo{year}{2017}\natexlab{}.
\newblock \showarticletitle{Google vizier: A service for black-box
  optimization}. In \bibinfo{booktitle}{\emph{Proceedings of the 23rd ACM
  SIGKDD international conference on knowledge discovery and data mining}}.
  \bibinfo{pages}{1487--1495}.
\newblock


\bibitem[Gramacy(2020)]%
        {gramacy2020surrogates}
\bibfield{author}{\bibinfo{person}{Robert~B Gramacy}.}
  \bibinfo{year}{2020}\natexlab{}.
\newblock \bibinfo{booktitle}{\emph{Surrogates: Gaussian process modeling,
  design, and optimization for the applied sciences}}.
\newblock \bibinfo{publisher}{CRC press}.
\newblock


\bibitem[Hennig and Schuler(2012)]%
        {hennig2012entropy}
\bibfield{author}{\bibinfo{person}{Philipp Hennig} {and}
  \bibinfo{person}{Christian~J Schuler}.} \bibinfo{year}{2012}\natexlab{}.
\newblock \showarticletitle{Entropy Search for Information-Efficient Global
  Optimization.}
\newblock \bibinfo{journal}{\emph{Journal of Machine Learning Research}}
  \bibinfo{volume}{13}, \bibinfo{number}{6} (\bibinfo{year}{2012}).
\newblock


\bibitem[Hern{\'a}ndez-Lobato et~al\mbox{.}(2014)]%
        {hernandez2014predictive}
\bibfield{author}{\bibinfo{person}{Jos{\'e}~Miguel Hern{\'a}ndez-Lobato},
  \bibinfo{person}{Matthew~W Hoffman}, {and} \bibinfo{person}{Zoubin
  Ghahramani}.} \bibinfo{year}{2014}\natexlab{}.
\newblock \showarticletitle{Predictive entropy search for efficient global
  optimization of black-box functions}.
\newblock \bibinfo{journal}{\emph{Advances in neural information processing
  systems}}  \bibinfo{volume}{27} (\bibinfo{year}{2014}).
\newblock


\bibitem[Hoffman et~al\mbox{.}(2011)]%
        {hoffman2011portfolio}
\bibfield{author}{\bibinfo{person}{Matthew Hoffman}, \bibinfo{person}{Eric
  Brochu}, \bibinfo{person}{Nando De~Freitas}, {et~al\mbox{.}}}
  \bibinfo{year}{2011}\natexlab{}.
\newblock \showarticletitle{Portfolio Allocation for Bayesian Optimization.}.
  In \bibinfo{booktitle}{\emph{UAI}}. \bibinfo{pages}{327--336}.
\newblock


\bibitem[Hofmann et~al\mbox{.}(2008)]%
        {hofmann2008kernel}
\bibfield{author}{\bibinfo{person}{Thomas Hofmann}, \bibinfo{person}{Bernhard
  Sch{\"o}lkopf}, {and} \bibinfo{person}{Alexander~J Smola}.}
  \bibinfo{year}{2008}\natexlab{}.
\newblock \showarticletitle{Kernel methods in machine learning}.
\newblock  (\bibinfo{year}{2008}).
\newblock


\bibitem[Hutter et~al\mbox{.}(2019)]%
        {hutter2019automated}
\bibfield{author}{\bibinfo{person}{Frank Hutter}, \bibinfo{person}{Lars
  Kotthoff}, {and} \bibinfo{person}{Joaquin Vanschoren}.}
  \bibinfo{year}{2019}\natexlab{}.
\newblock \bibinfo{booktitle}{\emph{Automated machine learning: methods,
  systems, challenges}}.
\newblock \bibinfo{publisher}{Springer Nature}.
\newblock


\bibitem[Hvarfner et~al\mbox{.}(2022)]%
        {hvarfner2022joint}
\bibfield{author}{\bibinfo{person}{Carl Hvarfner}, \bibinfo{person}{Frank
  Hutter}, {and} \bibinfo{person}{Luigi Nardi}.}
  \bibinfo{year}{2022}\natexlab{}.
\newblock \showarticletitle{Joint entropy search for maximally-informed
  Bayesian optimization}.
\newblock \bibinfo{journal}{\emph{arXiv preprint arXiv:2206.04771}}
  (\bibinfo{year}{2022}).
\newblock


\bibitem[Lam et~al\mbox{.}(2018)]%
        {lam2018advances}
\bibfield{author}{\bibinfo{person}{R{\'e}mi Lam}, \bibinfo{person}{Matthias
  Poloczek}, \bibinfo{person}{Peter Frazier}, {and} \bibinfo{person}{Karen~E
  Willcox}.} \bibinfo{year}{2018}\natexlab{}.
\newblock \showarticletitle{Advances in Bayesian optimization with applications
  in aerospace engineering}. In \bibinfo{booktitle}{\emph{2018 AIAA
  Non-Deterministic Approaches Conference}}. \bibinfo{pages}{1656}.
\newblock


\bibitem[Lindauer et~al\mbox{.}(2022)]%
        {lindauer2022smac3}
\bibfield{author}{\bibinfo{person}{Marius Lindauer}, \bibinfo{person}{Katharina
  Eggensperger}, \bibinfo{person}{Matthias Feurer}, \bibinfo{person}{Andr{\'e}
  Biedenkapp}, \bibinfo{person}{Difan Deng}, \bibinfo{person}{Carolin
  Benjamins}, \bibinfo{person}{Tim Ruhkopf}, \bibinfo{person}{Ren{\'e} Sass},
  {and} \bibinfo{person}{Frank Hutter}.} \bibinfo{year}{2022}\natexlab{}.
\newblock \showarticletitle{SMAC3: A Versatile Bayesian Optimization Package
  for Hyperparameter Optimization.}
\newblock \bibinfo{journal}{\emph{J. Mach. Learn. Res.}} \bibinfo{volume}{23},
  \bibinfo{number}{54} (\bibinfo{year}{2022}), \bibinfo{pages}{1--9}.
\newblock


\bibitem[Neiswanger et~al\mbox{.}(2022)]%
        {neiswanger2022generalizing}
\bibfield{author}{\bibinfo{person}{Willie Neiswanger}, \bibinfo{person}{Lantao
  Yu}, \bibinfo{person}{Shengjia Zhao}, \bibinfo{person}{Chenlin Meng}, {and}
  \bibinfo{person}{Stefano Ermon}.} \bibinfo{year}{2022}\natexlab{}.
\newblock \showarticletitle{Generalizing Bayesian Optimization with
  Decision-theoretic Entropies}.
\newblock \bibinfo{journal}{\emph{arXiv preprint arXiv:2210.01383}}
  (\bibinfo{year}{2022}).
\newblock


\bibitem[Nguyen et~al\mbox{.}(2020)]%
        {nguyen2020bayesian}
\bibfield{author}{\bibinfo{person}{Dang Nguyen}, \bibinfo{person}{Sunil Gupta},
  \bibinfo{person}{Santu Rana}, \bibinfo{person}{Alistair Shilton}, {and}
  \bibinfo{person}{Svetha Venkatesh}.} \bibinfo{year}{2020}\natexlab{}.
\newblock \showarticletitle{Bayesian optimization for categorical and
  category-specific continuous inputs}. In
  \bibinfo{booktitle}{\emph{Proceedings of the AAAI Conference on Artificial
  Intelligence}}, Vol.~\bibinfo{volume}{34}. \bibinfo{pages}{5256--5263}.
\newblock


\bibitem[Perego et~al\mbox{.}(2022)]%
        {perego2022autotinyml}
\bibfield{author}{\bibinfo{person}{Riccardo Perego}, \bibinfo{person}{Antonio
  Candelieri}, \bibinfo{person}{Francesco Archetti}, {and}
  \bibinfo{person}{Danilo Pau}.} \bibinfo{year}{2022}\natexlab{}.
\newblock \showarticletitle{AutoTinyML for microcontrollers: Dealing with
  black-box deployability}.
\newblock \bibinfo{journal}{\emph{Expert Systems with Applications}}
  \bibinfo{volume}{207} (\bibinfo{year}{2022}), \bibinfo{pages}{117876}.
\newblock


\bibitem[Regis(2016)]%
        {regis2016trust}
\bibfield{author}{\bibinfo{person}{Rommel~G Regis}.}
  \bibinfo{year}{2016}\natexlab{}.
\newblock \showarticletitle{Trust regions in Kriging-based optimization with
  expected improvement}.
\newblock \bibinfo{journal}{\emph{Engineering optimization}}
  \bibinfo{volume}{48}, \bibinfo{number}{6} (\bibinfo{year}{2016}),
  \bibinfo{pages}{1037--1059}.
\newblock


\bibitem[Ru et~al\mbox{.}(2020)]%
        {ru2020bayesian}
\bibfield{author}{\bibinfo{person}{Binxin Ru}, \bibinfo{person}{Ahsan Alvi},
  \bibinfo{person}{Vu Nguyen}, \bibinfo{person}{Michael~A Osborne}, {and}
  \bibinfo{person}{Stephen Roberts}.} \bibinfo{year}{2020}\natexlab{}.
\newblock \showarticletitle{Bayesian optimisation over multiple continuous and
  categorical inputs}. In \bibinfo{booktitle}{\emph{International Conference on
  Machine Learning}}. PMLR, \bibinfo{pages}{8276--8285}.
\newblock


\bibitem[Sch{\"o}lkopf et~al\mbox{.}(2002)]%
        {scholkopf2002learning}
\bibfield{author}{\bibinfo{person}{Bernhard Sch{\"o}lkopf},
  \bibinfo{person}{Alexander~J Smola}, \bibinfo{person}{Francis Bach},
  {et~al\mbox{.}}} \bibinfo{year}{2002}\natexlab{}.
\newblock \bibinfo{booktitle}{\emph{Learning with kernels: support vector
  machines, regularization, optimization, and beyond}}.
\newblock


\bibitem[Shahriari et~al\mbox{.}(2015)]%
        {shahriari2015taking}
\bibfield{author}{\bibinfo{person}{Bobak Shahriari}, \bibinfo{person}{Kevin
  Swersky}, \bibinfo{person}{Ziyu Wang}, \bibinfo{person}{Ryan~P Adams}, {and}
  \bibinfo{person}{Nando De~Freitas}.} \bibinfo{year}{2015}\natexlab{}.
\newblock \showarticletitle{Taking the human out of the loop: A review of
  Bayesian optimization}.
\newblock \bibinfo{journal}{\emph{Proc. IEEE}} \bibinfo{volume}{104},
  \bibinfo{number}{1} (\bibinfo{year}{2015}), \bibinfo{pages}{148--175}.
\newblock


\bibitem[Srinivas et~al\mbox{.}(2012)]%
        {srinivas2012information}
\bibfield{author}{\bibinfo{person}{Niranjan Srinivas}, \bibinfo{person}{Andreas
  Krause}, \bibinfo{person}{Sham~M Kakade}, {and} \bibinfo{person}{Matthias~W
  Seeger}.} \bibinfo{year}{2012}\natexlab{}.
\newblock \showarticletitle{Information-theoretic regret bounds for gaussian
  process optimization in the bandit setting}.
\newblock \bibinfo{journal}{\emph{IEEE transactions on information theory}}
  \bibinfo{volume}{58}, \bibinfo{number}{5} (\bibinfo{year}{2012}),
  \bibinfo{pages}{3250--3265}.
\newblock


\bibitem[Wang and Dowling(2022)]%
        {wang2022bayesian}
\bibfield{author}{\bibinfo{person}{Ke Wang} {and} \bibinfo{person}{Alexander~W
  Dowling}.} \bibinfo{year}{2022}\natexlab{}.
\newblock \showarticletitle{Bayesian optimization for chemical products and
  functional materials}.
\newblock \bibinfo{journal}{\emph{Current Opinion in Chemical Engineering}}
  \bibinfo{volume}{36} (\bibinfo{year}{2022}), \bibinfo{pages}{100728}.
\newblock


\bibitem[Wang et~al\mbox{.}(2021)]%
        {wang2021nextorch}
\bibfield{author}{\bibinfo{person}{Yifan Wang}, \bibinfo{person}{Tai-Ying
  Chen}, {and} \bibinfo{person}{Dionisios~G Vlachos}.}
  \bibinfo{year}{2021}\natexlab{}.
\newblock \showarticletitle{NEXTorch: a design and Bayesian optimization
  toolkit for chemical sciences and engineering}.
\newblock \bibinfo{journal}{\emph{Journal of Chemical Information and
  Modeling}} \bibinfo{volume}{61}, \bibinfo{number}{11} (\bibinfo{year}{2021}),
  \bibinfo{pages}{5312--5319}.
\newblock


\bibitem[Wang and Jegelka(2017)]%
        {wang2017max}
\bibfield{author}{\bibinfo{person}{Zi Wang} {and} \bibinfo{person}{Stefanie
  Jegelka}.} \bibinfo{year}{2017}\natexlab{}.
\newblock \showarticletitle{Max-value entropy search for efficient Bayesian
  optimization}. In \bibinfo{booktitle}{\emph{International Conference on
  Machine Learning}}. PMLR, \bibinfo{pages}{3627--3635}.
\newblock


\bibitem[Waring et~al\mbox{.}(2020)]%
        {waring2020automated}
\bibfield{author}{\bibinfo{person}{Jonathan Waring}, \bibinfo{person}{Charlotta
  Lindvall}, {and} \bibinfo{person}{Renato Umeton}.}
  \bibinfo{year}{2020}\natexlab{}.
\newblock \showarticletitle{Automated machine learning: Review of the
  state-of-the-art and opportunities for healthcare}.
\newblock \bibinfo{journal}{\emph{Artificial intelligence in medicine}}
  \bibinfo{volume}{104} (\bibinfo{year}{2020}), \bibinfo{pages}{101822}.
\newblock


\bibitem[Williams and Rasmussen(2006)]%
        {williams2006gaussian}
\bibfield{author}{\bibinfo{person}{Christopher~KI Williams} {and}
  \bibinfo{person}{Carl~Edward Rasmussen}.} \bibinfo{year}{2006}\natexlab{}.
\newblock \bibinfo{booktitle}{\emph{Gaussian processes for machine learning}}.
  Vol.~\bibinfo{volume}{2}.
\newblock \bibinfo{publisher}{MIT press Cambridge, MA}.
\newblock


\bibitem[Wilson et~al\mbox{.}(2020)]%
        {wilson2020efficiently}
\bibfield{author}{\bibinfo{person}{James Wilson}, \bibinfo{person}{Viacheslav
  Borovitskiy}, \bibinfo{person}{Alexander Terenin}, \bibinfo{person}{Peter
  Mostowsky}, {and} \bibinfo{person}{Marc Deisenroth}.}
  \bibinfo{year}{2020}\natexlab{}.
\newblock \showarticletitle{Efficiently sampling functions from Gaussian
  process posteriors}. In \bibinfo{booktitle}{\emph{International Conference on
  Machine Learning}}. PMLR, \bibinfo{pages}{10292--10302}.
\newblock


\bibitem[{\v{Z}}ilinskas and Calvin(2019)]%
        {vzilinskas2019bi}
\bibfield{author}{\bibinfo{person}{Antanas {\v{Z}}ilinskas} {and}
  \bibinfo{person}{James Calvin}.} \bibinfo{year}{2019}\natexlab{}.
\newblock \showarticletitle{Bi-objective decision making in global optimization
  based on statistical models}.
\newblock \bibinfo{journal}{\emph{Journal of Global Optimization}}
  \bibinfo{volume}{74} (\bibinfo{year}{2019}), \bibinfo{pages}{599--609}.
\newblock


\bibitem[Z{\"o}ller and Huber(2021)]%
        {zoller2021benchmark}
\bibfield{author}{\bibinfo{person}{Marc-Andr{\'e} Z{\"o}ller} {and}
  \bibinfo{person}{Marco~F Huber}.} \bibinfo{year}{2021}\natexlab{}.
\newblock \showarticletitle{Benchmark and survey of automated machine learning
  frameworks}.
\newblock \bibinfo{journal}{\emph{Journal of artificial intelligence research}}
   \bibinfo{volume}{70} (\bibinfo{year}{2021}), \bibinfo{pages}{409--472}.
\newblock


\end{thebibliography}


\newpage 
\appendix

\section{APPENDIX}

\begin{figure}[h!]
    \centering
    \includegraphics[width=0.6\textwidth]{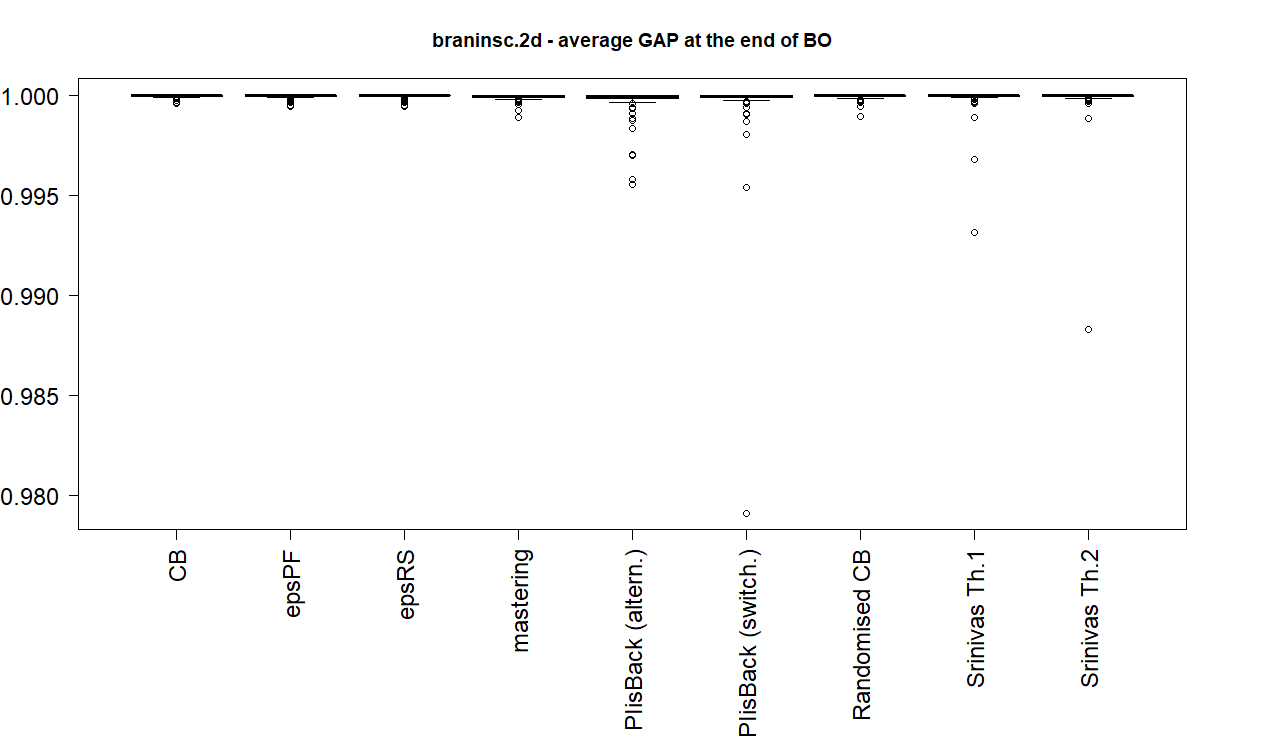}
    \caption{Branin test function: GAP values at the end of optimization processes (on 100 independent runs).}
    \label{fig:A1}
\end{figure}

\begin{figure}[h!]
    \centering
    \includegraphics[width=0.6\textwidth]{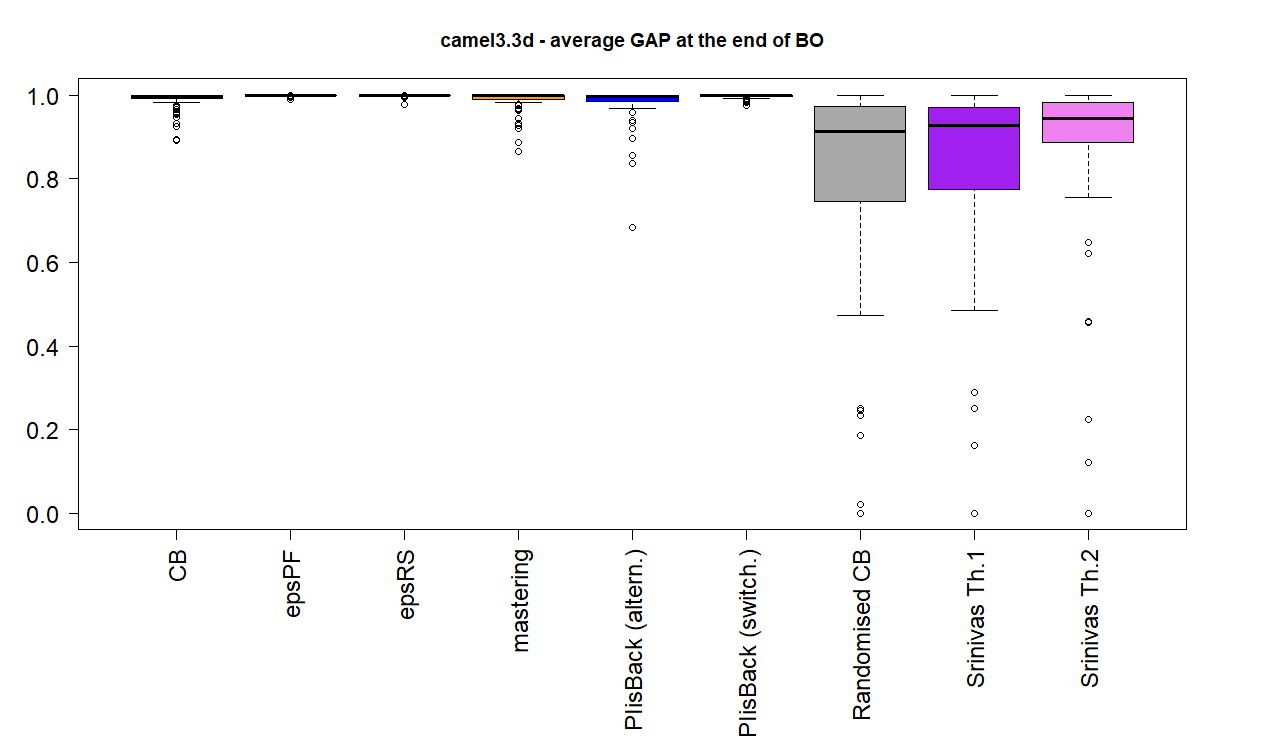}
    \caption{Camel3 test function: GAP values at the end of optimization processes (on 100 independent runs).}    
    \label{fig:A2}
\end{figure}

\begin{figure}[h!]
    \centering
    \includegraphics[width=0.6\textwidth]{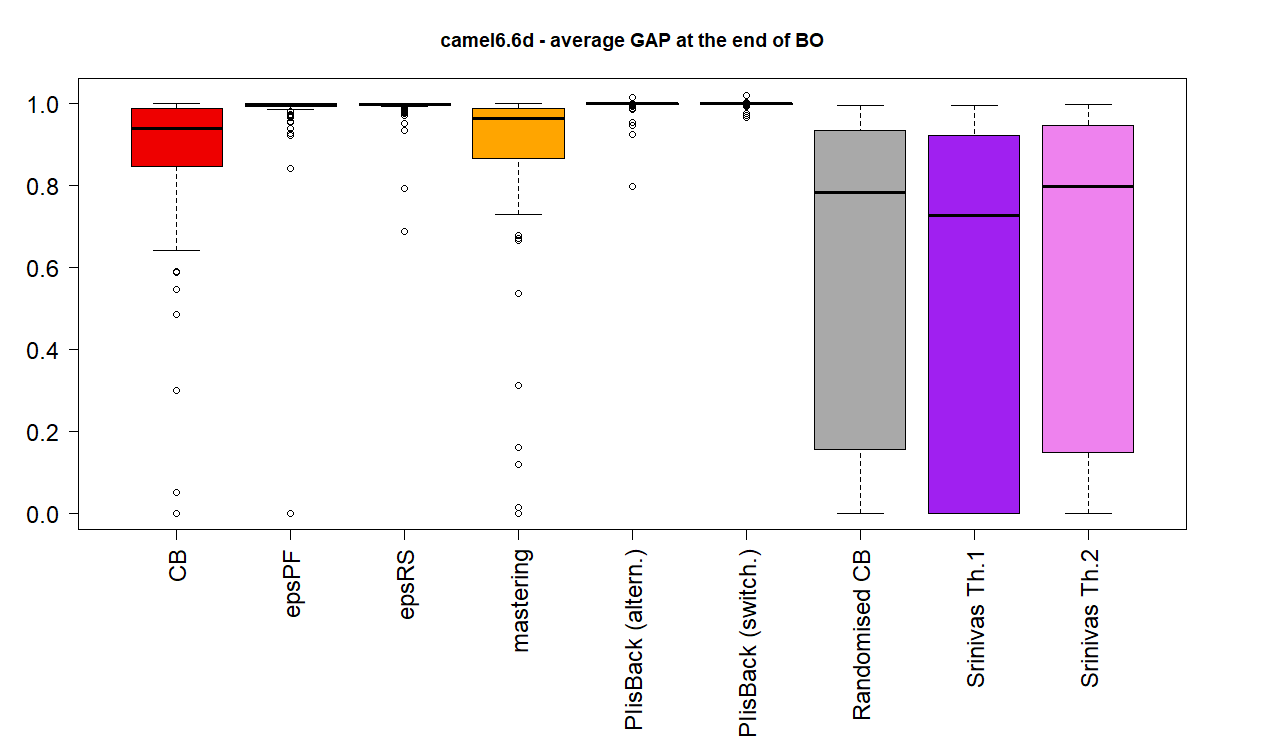}
    \caption{Camel6 test function: GAP values at the end of optimization processes (on 100 independent runs).}
    \label{fig:A3}
\end{figure}

\begin{figure}[h!]
    \centering
    \includegraphics[width=0.6\textwidth]{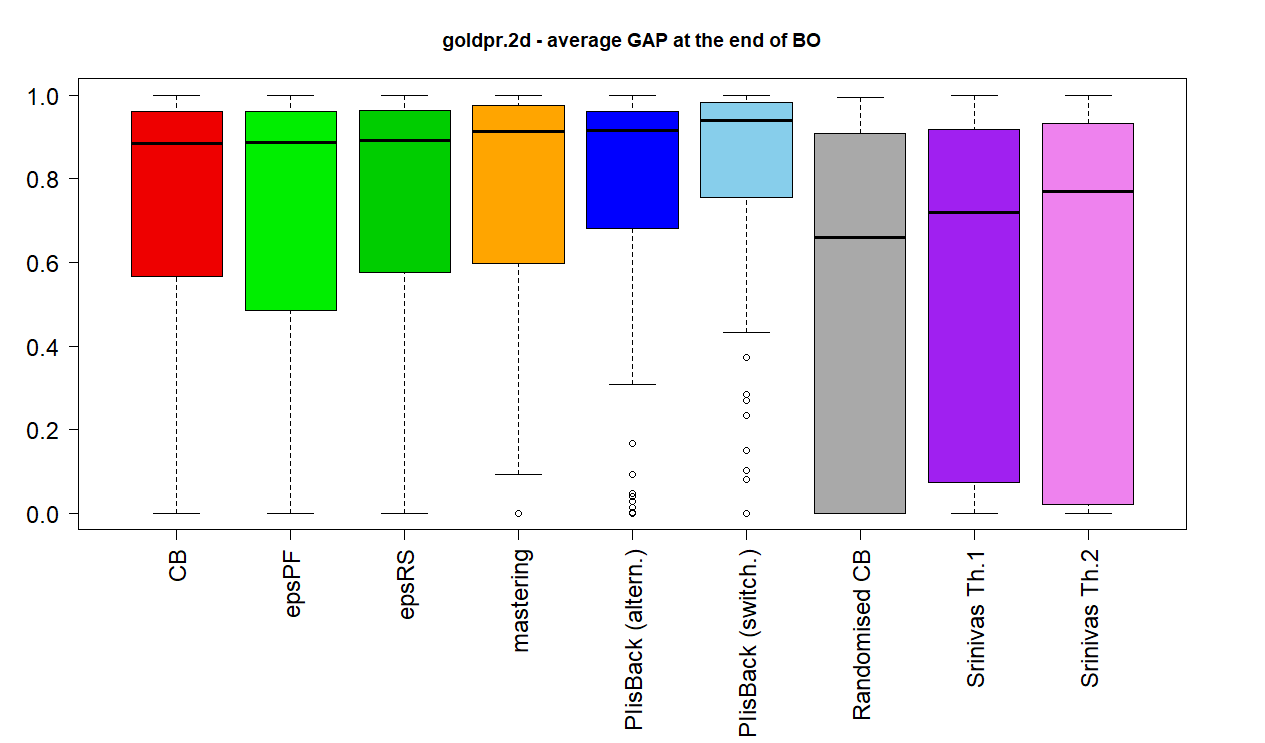}
    \caption{GoldPr test function: GAP values at the end of optimization processes (on 100 independent runs).}
    \label{fig:A4}
\end{figure}

\begin{figure}[h!]
    \centering
    \includegraphics[width=0.6\textwidth]{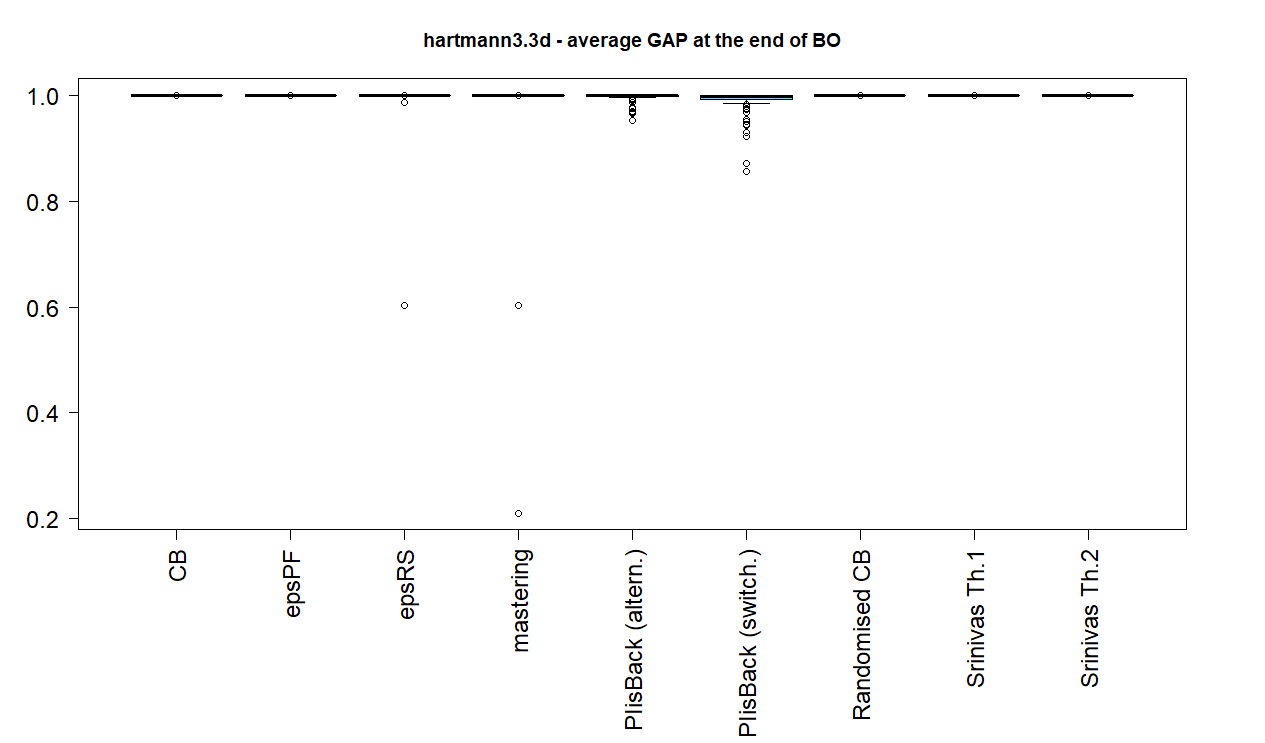}
    \caption{Hartmann3 test function: GAP values at the end of optimization processes (on 100 independent runs).}
    \label{fig:A5}
\end{figure}

\begin{figure}[h!]
    \centering
    \includegraphics[width=0.6\textwidth]{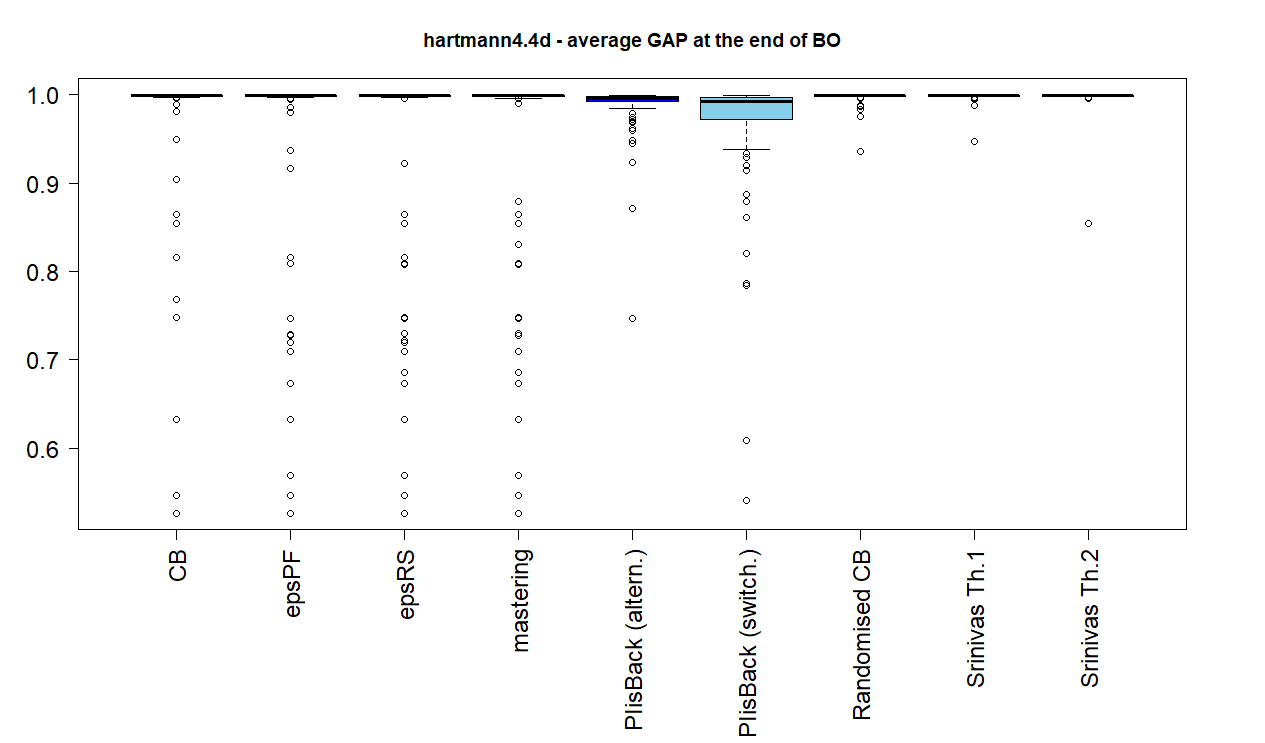}
    \caption{Hartmann4 test function: GAP values at the end of optimization processes (on 100 independent runs).}
    \label{fig:A6}
\end{figure}

\begin{figure}[h!]
    \centering
    \includegraphics[width=0.6\textwidth]{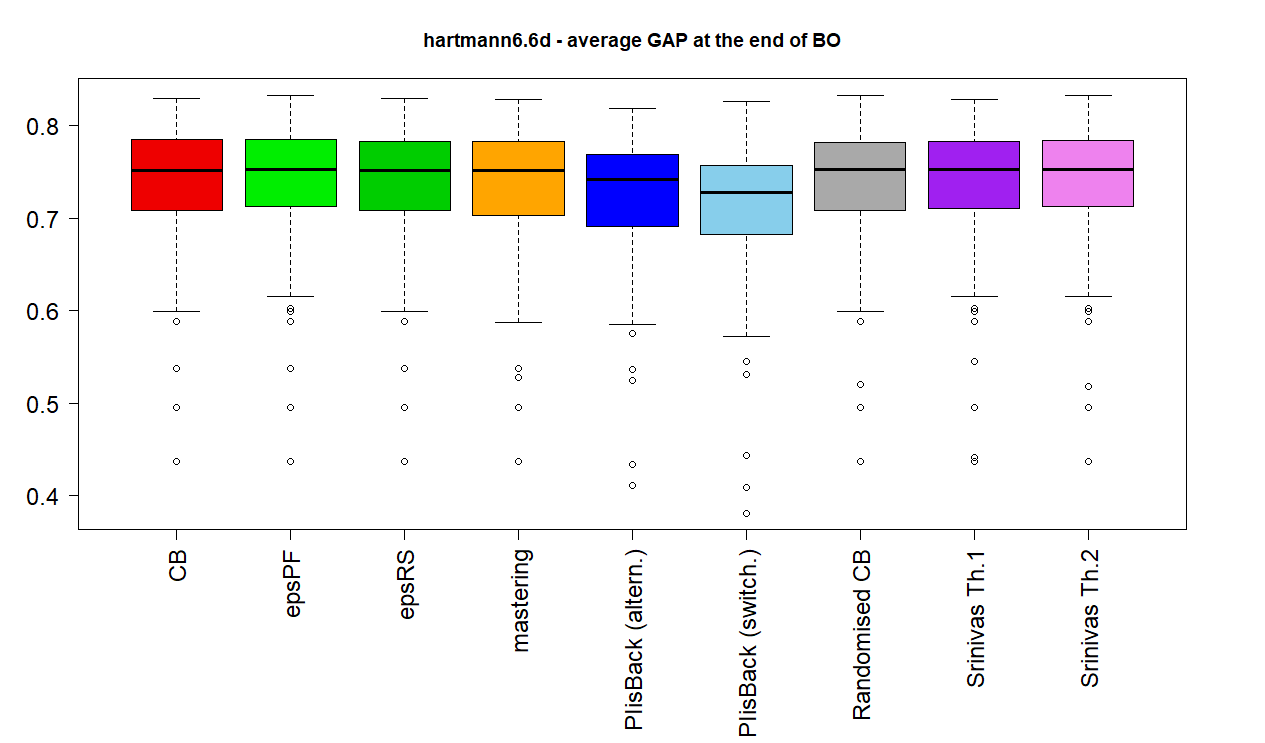}
    \caption{Hartmann6 test function: GAP values at the end of optimization processes (on 100 independent runs).}
    \label{fig:A7}
\end{figure}

\begin{figure}[h!]
    \centering
    \includegraphics[width=0.6\textwidth]{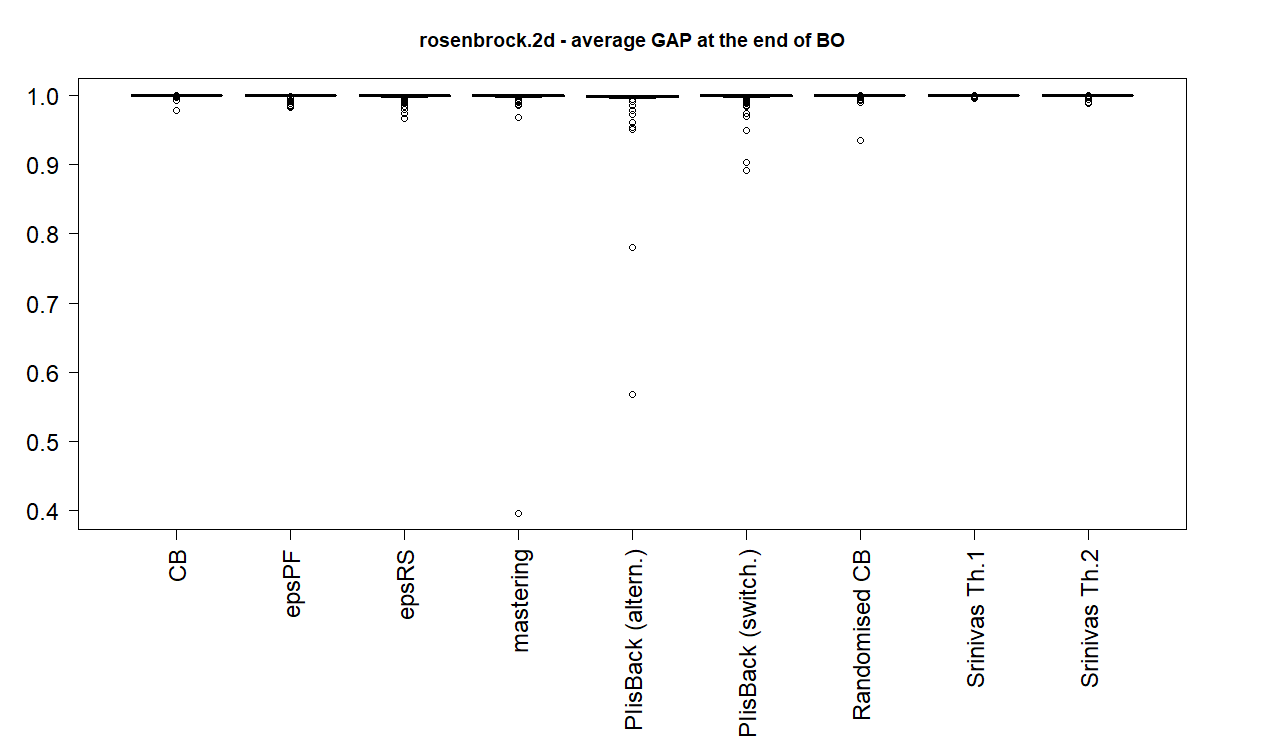}
    \caption{Rosenbrock test function: GAP values at the end of optimization processes (on 100 independent runs).}
    \label{fig:A8}
\end{figure}

\begin{figure}[h!]
    \centering
    \includegraphics[width=0.6\textwidth]{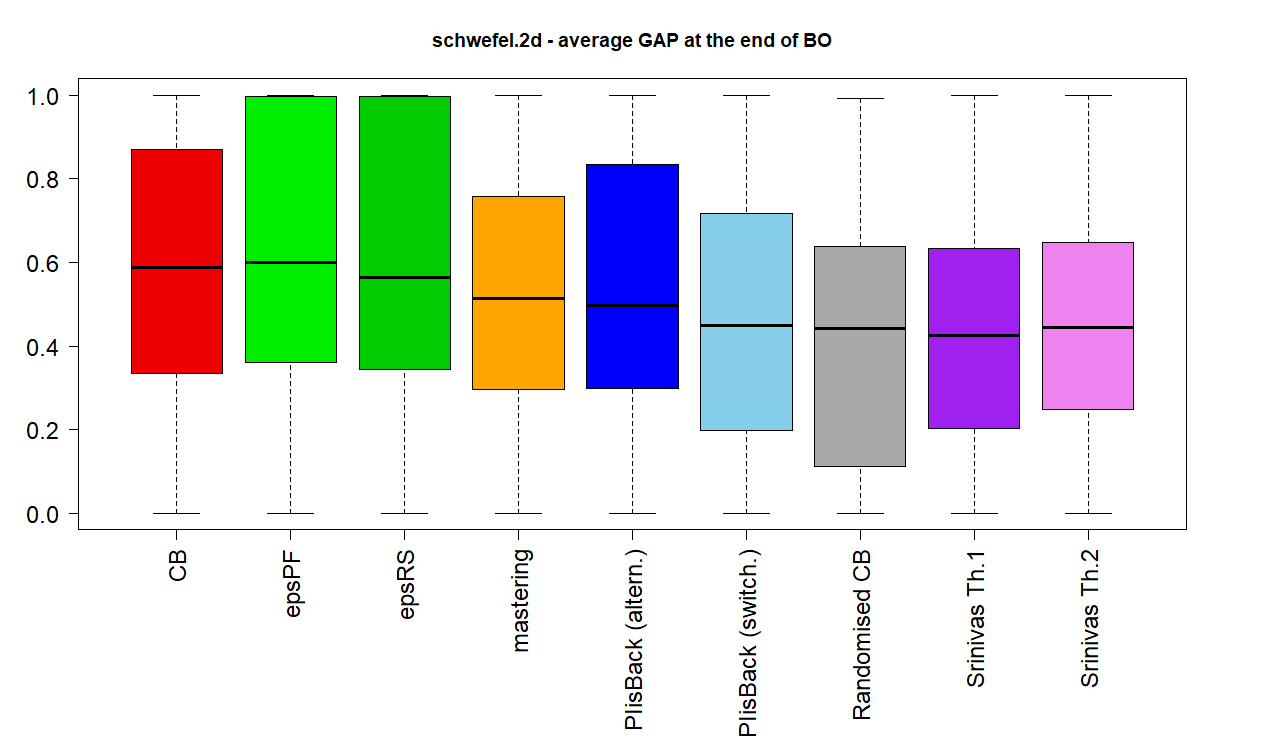}
    \caption{Schwefel test function: GAP values at the end of optimization processes (on 100 independent runs).}    
    \label{fig:A9}
\end{figure}

\begin{figure}[h!]
    \centering
    \includegraphics[width=0.6\textwidth]{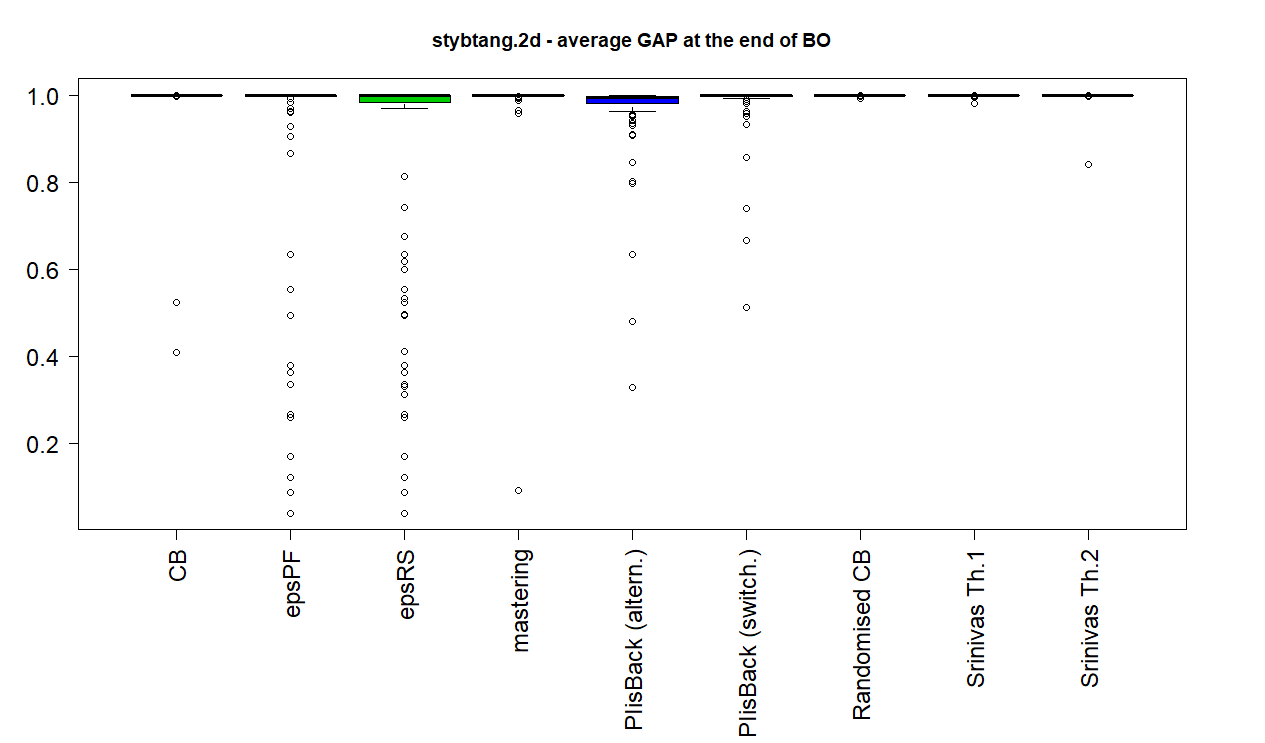}
    \caption{StybTang test function: GAP values at the end of optimization processes (on 100 independent runs).}
    \label{fig:A10}
\end{figure}

\end{document}